\documentclass[a4paper]{article}

\usepackage[english]{babel}
\usepackage[utf8x]{inputenc}
\usepackage[T1]{fontenc}

\usepackage[a4paper,top=3cm,bottom=2cm,left=3cm,right=3cm,marginparwidth=1.75cm]{geometry}

\usepackage{amsmath,amsthm}
\usepackage{amssymb,mathrsfs}
\usepackage{mathtools}
\usepackage{bm}
\usepackage{graphicx}
\usepackage{wrapfig}
\usepackage{subcaption}

\graphicspath{ {images/} }

\DeclareMathOperator*{\argmax}{\arg\!\max}

\title{Adversary Detection in Neural Networks via Persistent Homology}
\author{
  Thomas Gebhart\footnote{email: gebha095@umn.edu}\\
  \textit{\footnotesize{University of Minnesota}}
  \and
  Paul Schrater\\
  \textit{\footnotesize{University of Minnesota}}
}
\date{}
\begin{document}
\maketitle

\begin{abstract}

We outline a detection method for adversarial inputs to deep neural networks. By viewing neural network computations as graphs upon which information flows from input space to output distribution, we compare the differences in graphs induced by different inputs. Specifically, by applying persistent homology to these induced graphs, we observe that the structure of the most persistent subgraphs which generate the first homology group differ between adversarial and unperturbed inputs. Based on this observation, we build a detection algorithm that depends only on the topological information extracted during training. We test our algorithm on MNIST and achieve 98\% detection adversary accuracy with $\text{F}_1$-score 0.98. 

\end{abstract}

\section{Introduction}

Artificial neural networks (neural networks) are a class of machine learning algorithms that learn to accomplish tasks by considering example data and iteratively reconfiguring themselves. This ability to learn input representations without task-specific programming has proven to be extremely powerful in solving machine learning problems across a multitude of domains. For example, neural networks have been used successfully in image and speech recognition \cite{szegedy2015going,krizhevsky2012imagenet,farabet2013learning,tompson2014joint,mikolov2011strategies,hinton2012deep} as well as in the reconstruction of brain circuits \cite{helmstaedter2013connectomic}, analysis of particle accelerator data \cite{ciodaro2012online}, prediction of gene mutation \cite{leung2014deep}, and language translation \cite{sutskever2014sequence} among many other applications. 

Much research has been done to better understand these neural network computations. For example, Li et al. \cite{li2015convergent} look at bipartite matching of neuron activations to determine whether different neural networks learn the same representations as encoded by these neural activations. They determine that neural network architectures trained on the same data generally do encode similar representations. As well, tools for visualizations of hidden neural network layers have shown, in a qualitative sense, how neural networks might build up these complex representations of their input \cite{yosinski2015understanding,zeiler2014visualizing}. These layer-wise visualizations imply that neural networks build representations from simple, Gabor-like features in early layers into more complex features like faces, circles, or flower petals in later layers. These results seem to imply that neural networks are able to robustly capture the relevant semantic information within their input in an intuitive manner. 

However, Szegedy et al. \cite{szegedy2013intriguing} showed that by adding human-imperceptible noise to an input example, neural networks can be fooled into misclassifying this now \textit{adversarial example} that it originally correctly classified, despite very little change in the input. These adversarial examples have been shown to exist in numerous domains across many network architectures \cite{moosavi2016universal}. 

Much recent work has focused on correctly classifying these adversarial examples \cite{huang2015learning, gu2014towards, bastani2016measuring}. Unfortunately, these methods tend to perform poorly as adversarial examples are deeply linked to the nonlinear computations and deformations of input space that provide much of the power derived from neural networks. Noting this, researchers have turned towards the detection of adversarial examples outright \cite{bhagoji2017dimensionality, metzen2017detecting, li2016adversarial}. Carlini et al. \cite{carlini2017adversarial}, in an excellent review, show that these detection methods are also not particularly robust, especially in scenarios where the adversary has information on the detection algorithm.

The existence of adversarial examples in neural networks reveals that the internal representations neural networks use to make classification decisions may not be as robust as intuition expects. The ubiquity of adversarial examples across neural network architectures and between problem domains points to the understanding that adversarial examples exploit the brittle, nonlinear representation structure that neural networks employ in classification. Given this, it is natural to ask whether this exploitation is detectable within the structure of the computation itself. 

Motivated by this question, we construct a framework for quantifying, in a global sense, neural network computations for a given input. By viewing neural network computations as graphs, we can formally trace the flow of information from the input layer to the output layer. We look at the persistent homology of the graphs induced by different inputs and view the difference in their persistent subgraphs. This formalization allows us to quantify how different any two classification decisions are in terms of connected components and flow of information through the input-induced computational graph. With this construction, we investigate how adversarial examples are reflected in changes in the induced computational graph. 

The outline of the paper is as follows. Section 2 formally introduces neural network computations and explains how they may be viewed as weighted, directed graphs. Section 3 introduces persistent homology and how it relates to graphs of neural network computations. Section 4 describes the adversaries studied in the paper. Section 5 summarizes the results of the paper and gives recommendations of future work. Section 6 offers concluding remarks. 

\section{Neural Networks as Graphs}\label{sec:networkgraph}

The results of this paper rely on the ability to represent neural networks as computational graphs. In general, we look to model information flow through a neural network from input to probability distribution. This idea is well-founded given the close relationship of convolutional, recurrent, and fully-forward neural networks to deep belief and Bayesian networks. This section outlines the graphical construction for convolutional and fully-connected layers because we use only these layers in the tests later in the paper. However, other types of neural network layers, like pooling or recurrent layers, can be realized under the same paradigm. 

There exist multiple ways to view neural networks as directed graphs of information flow. The first is a static representation in which the network's filters and weight matrices define the potential for information flow through the network. The second is the realized information flow through the network after an input is added. This paper focuses on this second representation, as we are interested in differences in information flow across different inputs. 

\subsection{Fully-Connected Layers}

Fully-connected layers are structured such that every neuron in the output of the previous layer is connected to each neuron in the next layer. This computational structure is typically implemented as matrix multiplication where the left matrix has elements consisting of each neuron's activation value in the previous layer's output, and the right matrix contains the weighted connections between the previous layer's output and the next layer's output neurons. 


The graphical representation of a fully-connected layer then naturally simplifies to a fully-connected graph with weights described by the resultant multiplicative values of the weight matrix being multiplied by the previous layer's activations. These multiplicative values are induced when an input is fed into the network and the activations propagate to the layer's weight matrix. The vertices of a fully-connected layer's graphical representation are the previous layer's output neurons along with the next layer's output neurons. At the output activation propagation to the next layer, the neurons perform an addition operation on all of the input edges (weights) to determine the input to its activation function then passes equally the resulting value as its activation value.

\subsection{Convolutional Layers}

Convolutional layers are characterized by performing a convolution operation of output activations from the previous layer with a filter matrix. Again, one can view this layer's operation in terms of matrix multiplication. Namely, we move the filter in a predetermined stride across the matrix formed by the output activations of the previous layer and compute a matrix multiplication operation at each stride step. The output of this operation determines the activation values of the output neurons.

\begin{figure}[!htb]
\caption{A convolution matrix multiplication operation (left) and its graphical representation (right).}\label{fig:graph}
\minipage{\textwidth}
	\includegraphics[width=0.58\linewidth, keepaspectratio]{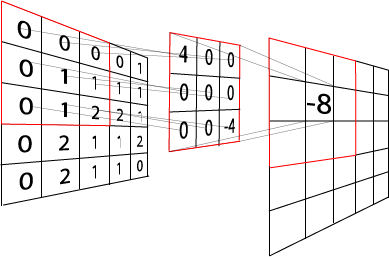}
	\includegraphics[width=0.4\linewidth, keepaspectratio]{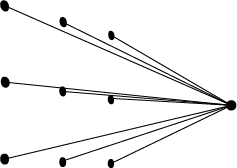}
\endminipage
\end{figure}

The graphical view of this operation can be realized in a similar manner to the fully-connected case. The connections between the output neurons of the previous layer to the output neurons of the next layer are determined largely by the size of the filter, the stride, and the filter values. We can visualize these edge connections by viewing the pre-image of the filter matrix at each stride on the previous layer's output neurons such that each neuron in the pre-image connects to a single neuron in the next output layer with edge weight equal to the activation value of that neuron multiplied by the filter value at that location in the image. The vertices of this graph are the output neurons from the previous layer and the output neurons of the next layer. An output neuron of this layer performs an addition operation on all of its input edge weights to determine its input to its activation function then passes equally the resulting value as its activation value.

\section{Persistent Homology}

With a proper representation of our neural network as a directed graph, we can now view differences in the graphs induced by different inputs. One way to measure these differences is via persistent homology. Persistent homology is a tool from topological data analysis that allows one to summarize the global topological information of a discretely approximated space. Persistent homology counts the number of ``holes" in different dimensions of a space at different thresholds of connectedness within the underlying points. More persistent features (holes) that survive over a larger variation of threshold values are more likely to represent actual features of the space and are less likely to be produced by noise. There exist numerous open-source libraries that include functionality for computing persistent homology \cite{morozov2017dionysus,nanda2012perseus,henselman2016matroid}. Persistent homology is defined on simplicial complexes, of which weighted graphs are an example. 

More formally, consider a real-valued function on a simplicial complex $f: K \rightarrow \mathbb{R}$ that is non-decreasing on increasing sequences of faces $f(\sigma) < f(\tau)$ when $\sigma$ is a face of $\tau$ in $K$. For all $q \in \mathbb{R}$ the sublevel set $K(q) = f^{-1}(-\infty, q]$ is a subcomplex of $K$. The ordering of the values of $f$ on $K$ induces an ordering on the sublevel complexes. We then have a filtration $\emptyset = K_0 \subseteq K_1 \subseteq \dots \subseteq K_n = K$. When $i \leq j \leq n$ the inclusion $K_i \xhookrightarrow{} K_j$ induces a homomorphism $f_p^{i,j}: H_p(K_i) \rightarrow H_p(K_j)$ on the $p$th simplicial homology group where $p$ is the dimension. The $p$th persistent homology groups are the images of these homomorphisms and the $p$th persistent Betti numbers $\beta_p^{i,j}$ are the associated ranks of these groups. A persistence module over a partially ordered set $V$ is a set of vector spaces $U_t$ indexed by $V$ with associated linear maps $u_t^s: U_s \rightarrow U_t$ when $s \leq t$. We can consider this function a functor from $V$ considered as a category to the category of vector spaces. We then have a classification of persistence modules over a field $F$ indexed by $\mathbb{N}$: 
\begin{equation*}
	U \simeq \oplus_i x^{t_i} \cdot F[x] \oplus (\oplus_j x^{r_j} \cdot (F[x]/(x^{s_j} \cdot F[x]))) 
\end{equation*}

Multiplication by $x$ moves up one step in the persistence module. The free parts of the right side of the equation correspond to the homology generators that appear (are born) at filtration level $t_i$ and do not disappear. The torsion elements correspond to homology generators that are born at $r_j$ and disappear (die) at $s_j$ in the filtration. Using this information, we may uniquely represent the persistent homology of a simplicial complex as a persistence diagram which plots a point for each generator with its x-coordinate corresponding to its birth time and its y-coordinate the death time.

Given two persistence diagrams $X$ and $D$, we can define the Wasserstein distance between them: 

\begin{equation} \label{eq: wasserstein}
	W_p(X,D) = \inf_{\phi: X \rightarrow D} \left(\sum\limits_{x \in X} \| x - \phi(x) \|_q\right)^{\frac{1}{p}}
\end{equation}

where $p \geq 1$, $q \leq \infty$, and $\phi$ ranges over bijections between $X$ and $D$. Letting $p \rightarrow \infty$ gives the bottleneck distance. In this paper, we take $p = 2$. One can view the Wasserstein distance on persistence diagrams as a minimal bipartite matching of points between the diagrams where points are matched by minimum distance in the birth-death plane.

\subsection{Persistent Homology on Neural Networks}

The Wasserstein distance between persistence diagrams gives us a way to measure topological similarity between graphs induced by different inputs. For this to work with our characterization of induced neural network graphs, we need to make a few changes. 

First, we take the absolute value of edge weights in our induced neural network graph. The filtration step of the persistent homology calculation works on a sorted list of simplices by value. In our framework of information flow, large negative edge weights are as semantically relevant to the characterization of our network as large positive edge weights because both have a large impact on the next layer. 

Second, we sort simplices in the 0th and 1st dimensions from largest to smallest which is the reverse of what is typically implemented in persistent homology calculations. This sorting is more applicable for our usage given that larger edge weights better characterize the global information flow through the network than smaller edge weights and should thus appear first in the filtration. 

Finally, we make a small augmentation to the calculation of the Wasserstein distance between the persistence diagrams of our input-induced graphs. Components in graphs that persist through the last filtration value (maximum edge value in our case) are typically assigned a death time of infinity in the persistence calculation. At each dimension, there will always be at least one point at infinity. Because our filtration goes from largest edge weight to smallest, components that do not merge with other simplices and die are given death time equal to the lowest edge weight in the filtration.

In this paper, we focus on the the zeroth homology group $H_0$, the persistent homology of which corresponds to the life-death pairs of connected components within a neural network graph induced by an input. It is possible in a feed-forward architecture like a convolutional neural network to have non-trivial $H_1$, but the intuition as to what these layer-skipping holes mean in terms of network semantics is slightly more opaque. As well, because we truncate the death times of non-dying components, the death times of these $H_1$ holes are all shared which makes computing the Wasserstein distance more computationally expensive.

\subsection{Software}

We use the second version of the C++ library Dionysus \cite{morozov2017dionysus} for the computation of persistent homology. We use Tensorflow \cite{abadi2016tensorflow} for neural network construction and computations. For ease of use, we implement the computations found in the Dionysus library as a custom operation within the Tensorflow library. We also make the changes described in the previous section in this custom Tensorflow operation\footnote{This code is available at github.com/tgebhart/dionysus\_tensorflow.}. The analysis in the Results section is performed primarily in Python\footnote{The code used in this analysis can be found at github.com/tgebhart/tf\_activation.}.

\section{Adversarial Examples}\label{sec:adversaries}

We generate adversarial images for MNIST \cite{lecun1998gradient} as described in \cite{carlini2017towards}. As far as the authors know, this is the strongest method for generating adversarial examples in that the adversaries generated are able to fool many adversary detection schemes outright and can be augmented to fool others by altering the loss function \cite{carlini2017adversarial}. We give a brief summary of the adversary generation algorithm below, but a more in-depth treatment can be found in \cite{carlini2017towards}. 

\subsection{Neural Network Notation}

A neural network may be simplified as a function $F:  \mathbb{R}^n \rightarrow \mathbb{R}^m$ that accepts input $\bm{x} \in \mathbb{R}^n$ and maps this to a class $\bm{y} \in \mathbb{R}^m$. Our input in this paper is a two-dimensional $l \times w = n$ pixel grey-scale image where $x_i \in \bm{x}$ is the intensity of pixel $i$. For MNIST, this corresponds to grey-scale images of size $28 \times 28$ which we vectorize so that $\bm{x} \in \mathbb{R}^{784}$. 

The output of the network is computed using a softmax function which forces each element $y_i$ of $\bm{y}$ (the logits) to the range $0 \leq y_i \leq 1$ such that 

\begin{equation*}
\sum\limits_{i = 1}^m y_i = 1
\end{equation*}.

Thus, we can view the output of neural networks as a probability distribution over all classes $y_i \in \bm{y}$. The predicted class assignment $C(\bm{x})$ from the neural network is assigned via $C(\bm{x}) = \argmax_i F(\bm{x})_i$. We define the correct class label as $C^*(\bm{x})$ for input example $\bm{x}$. Combining this information, our final output from the neural network can be characterized as

\begin{equation*}
F(\bm{x}) = \text{softmax}(Z(\bm{x})) = \bm{y}
\end{equation*}

where $Z(\bm{x}) = \bm{z}$ are the output logits before softmax. 

\subsection{Adversarial Examples}

\begin{figure}
\begin{subfigure}{\textwidth}
  \includegraphics[width=.09\linewidth]{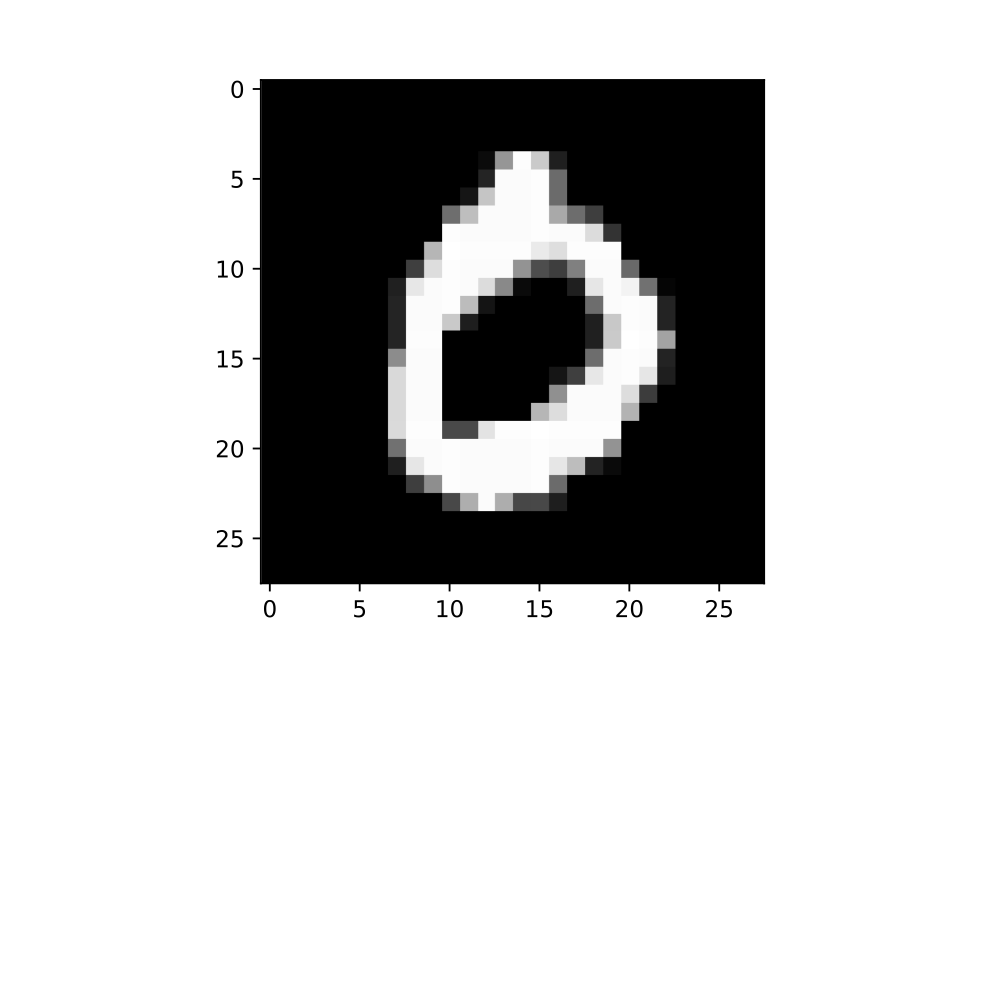}
  \includegraphics[width=.09\linewidth]{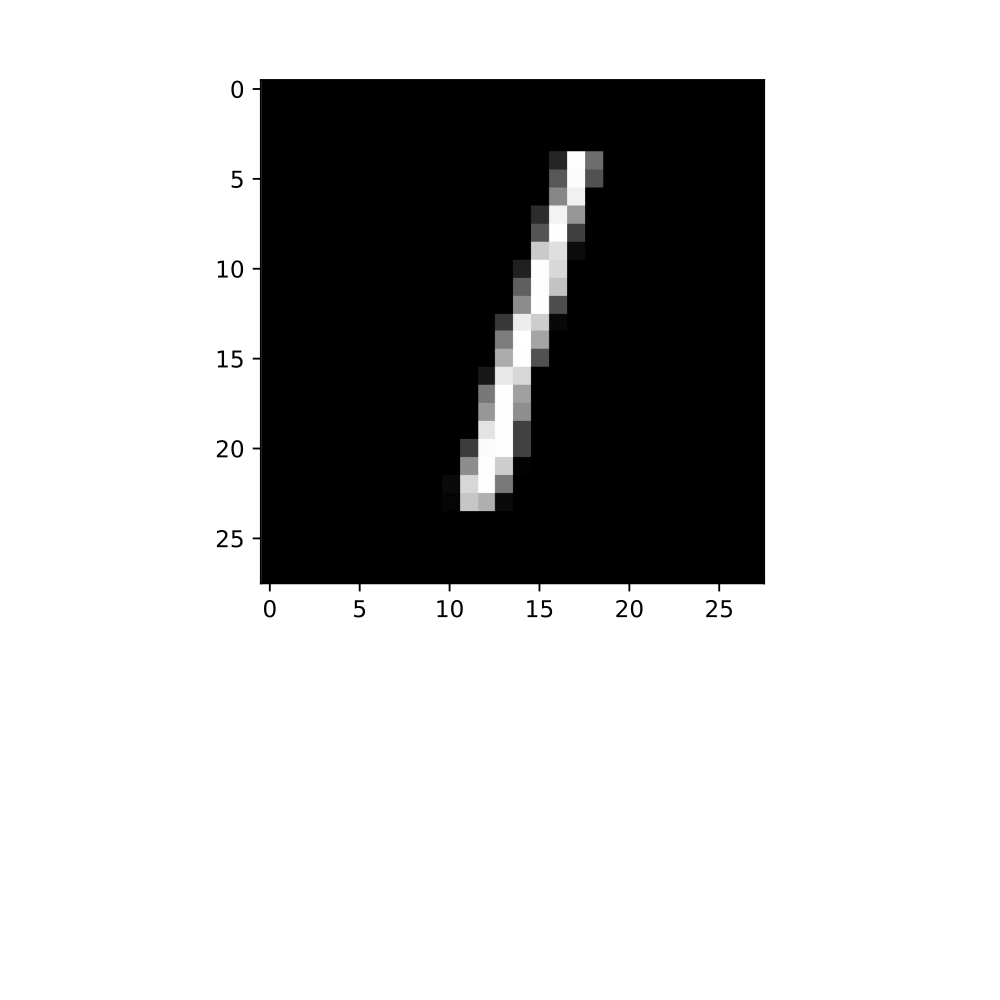}
  \includegraphics[width=.09\linewidth]{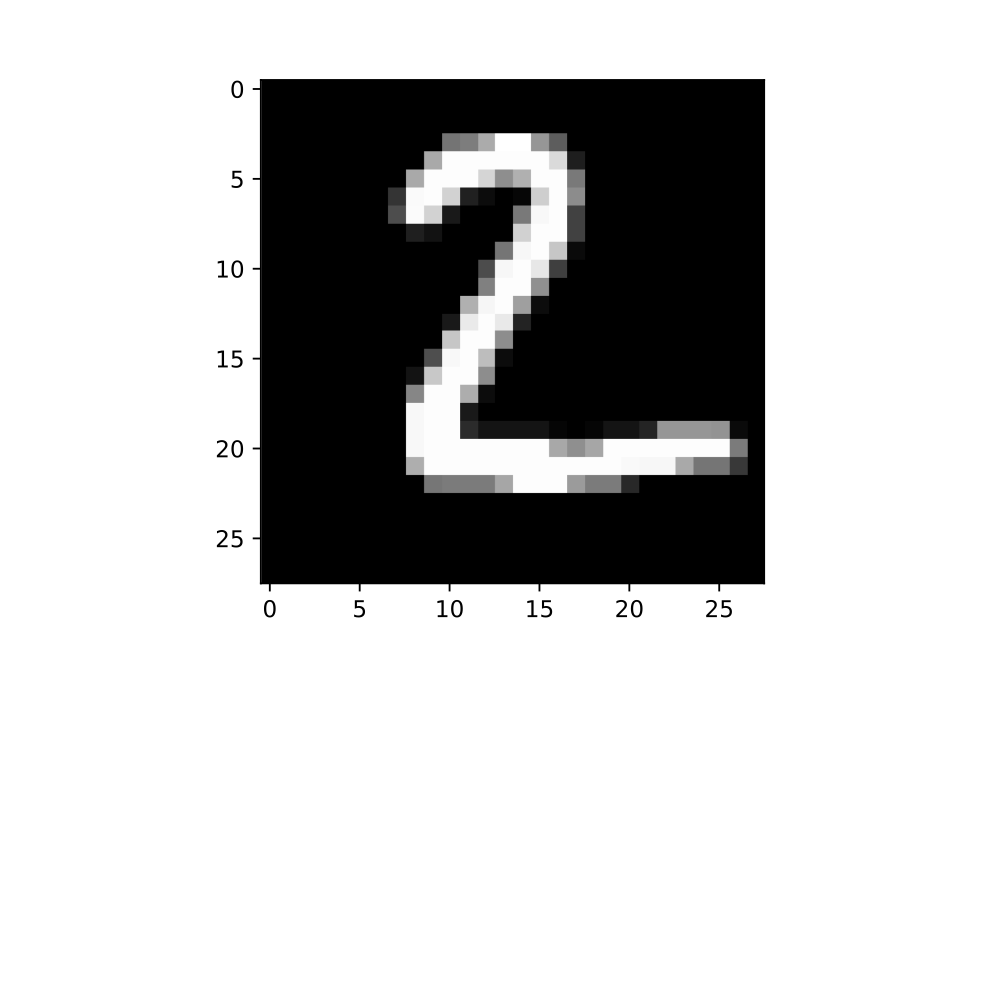}
  \includegraphics[width=.09\linewidth]{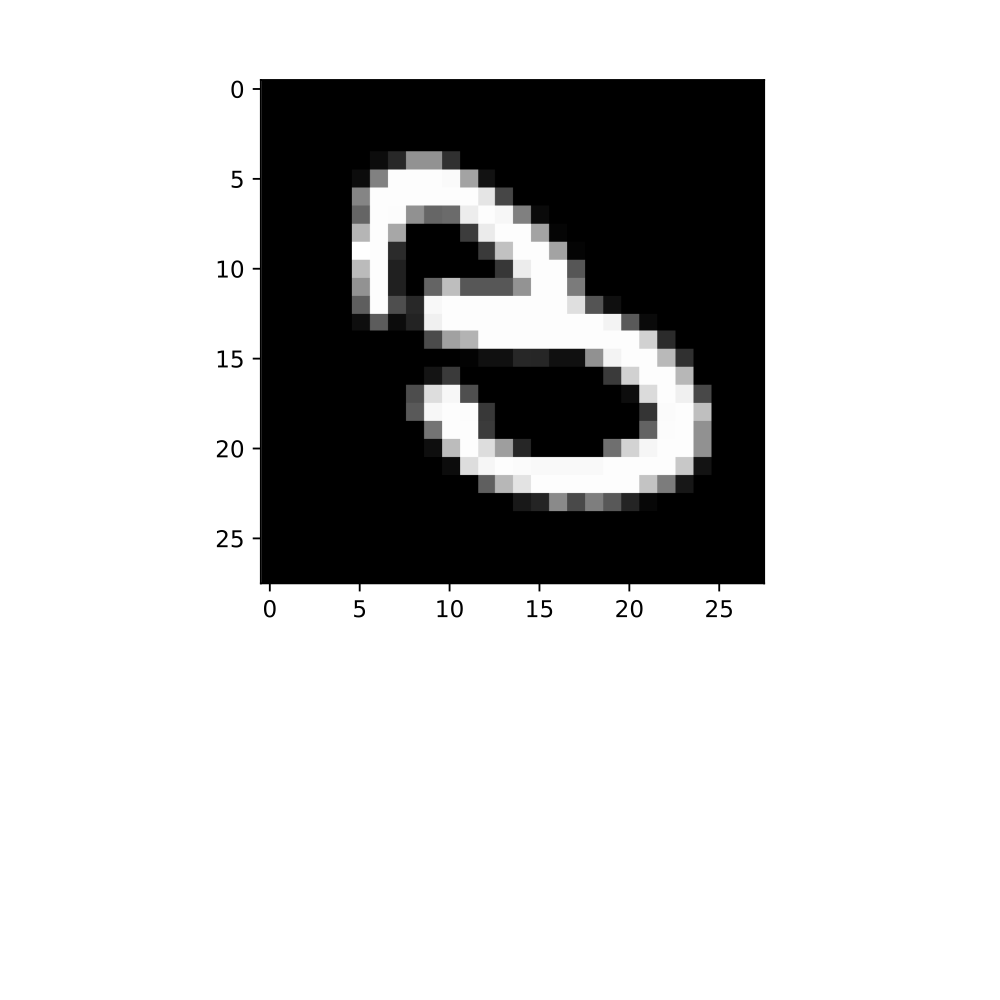}
  \includegraphics[width=.09\linewidth]{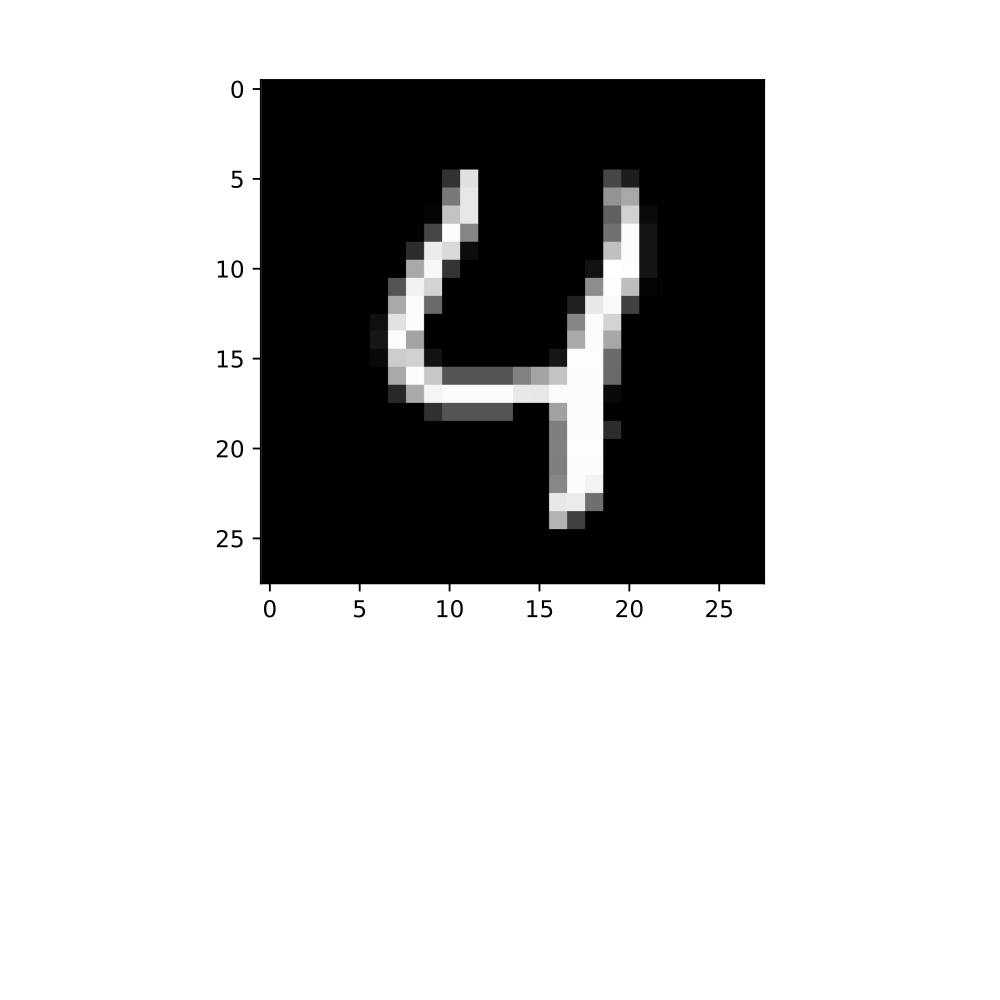}
  \includegraphics[width=.09\linewidth]{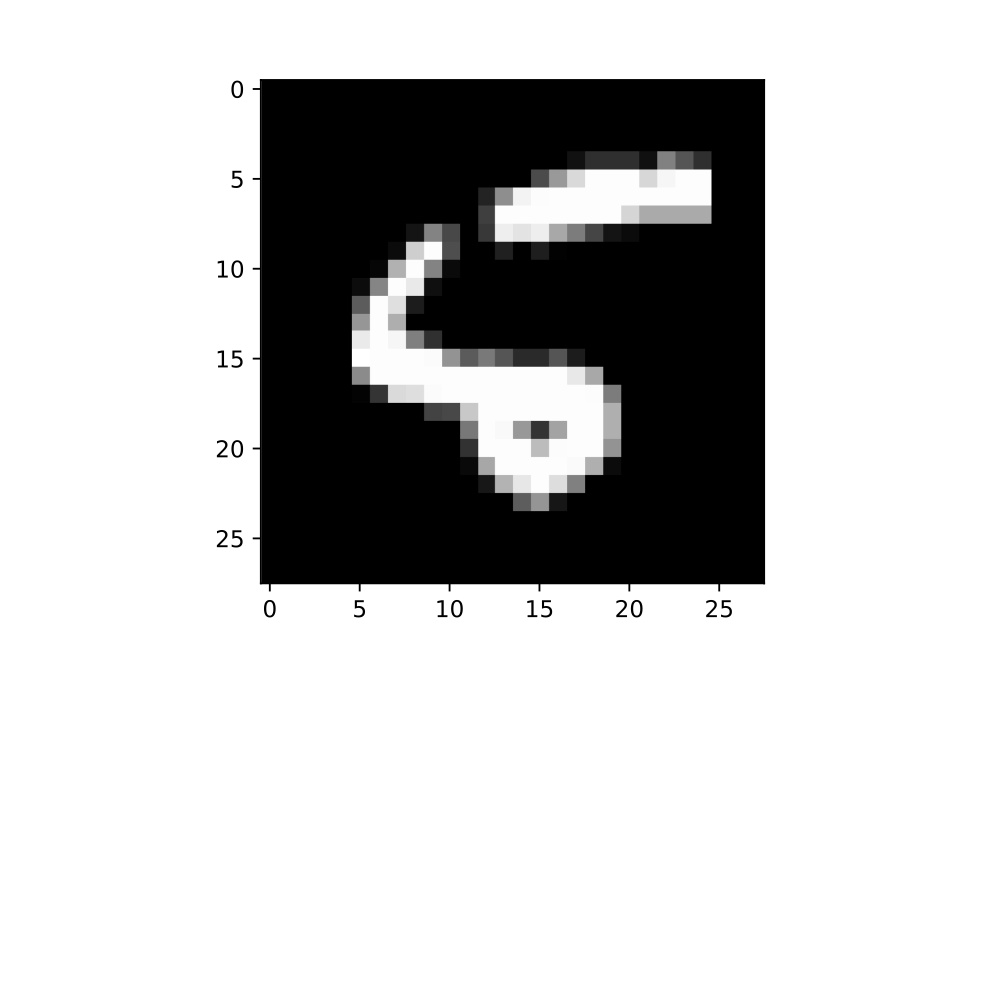}
  \includegraphics[width=.09\linewidth]{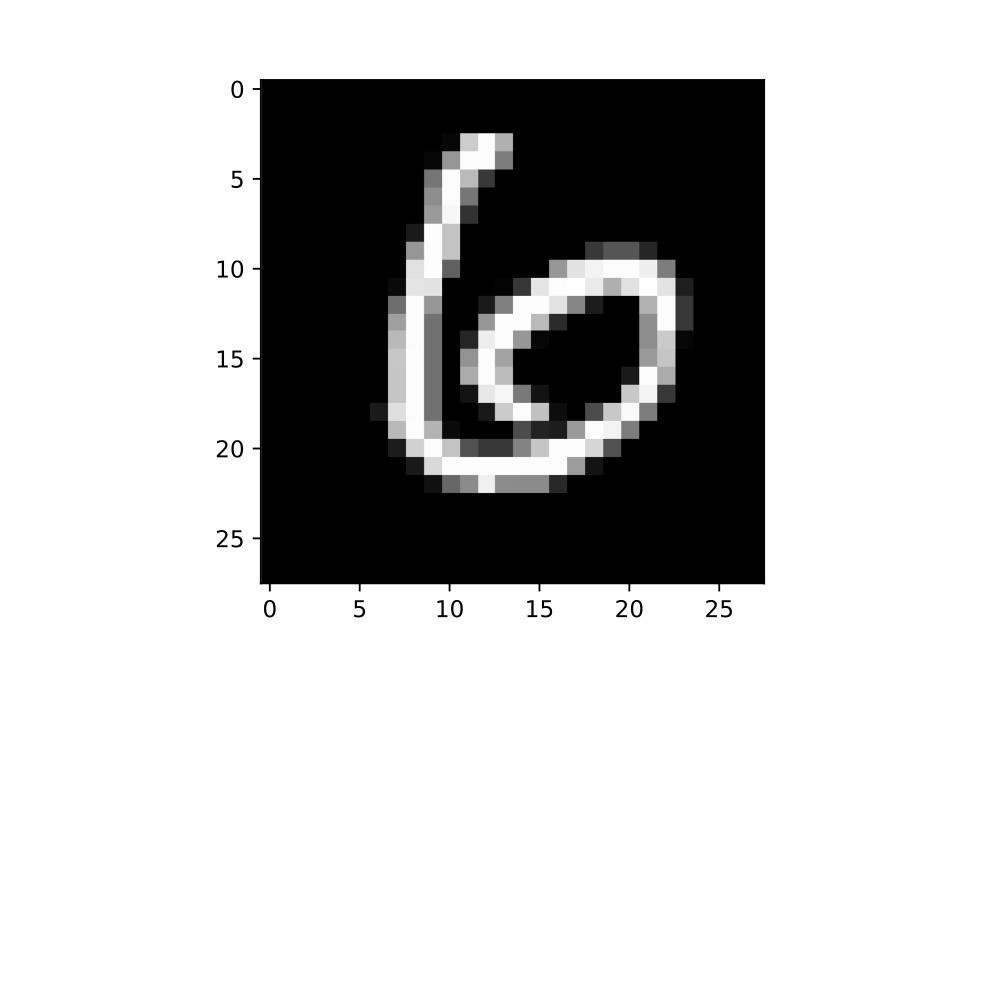}
  \includegraphics[width=.09\linewidth]{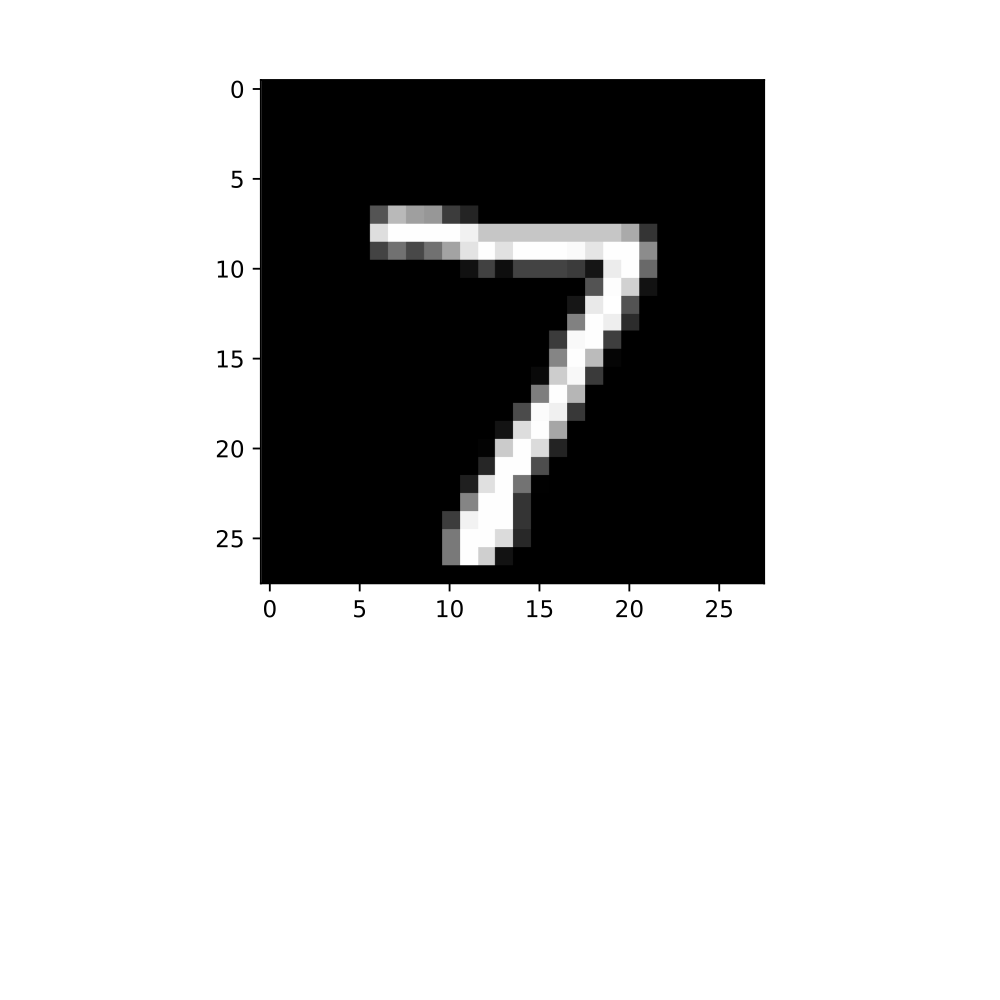}
  \includegraphics[width=.09\linewidth]{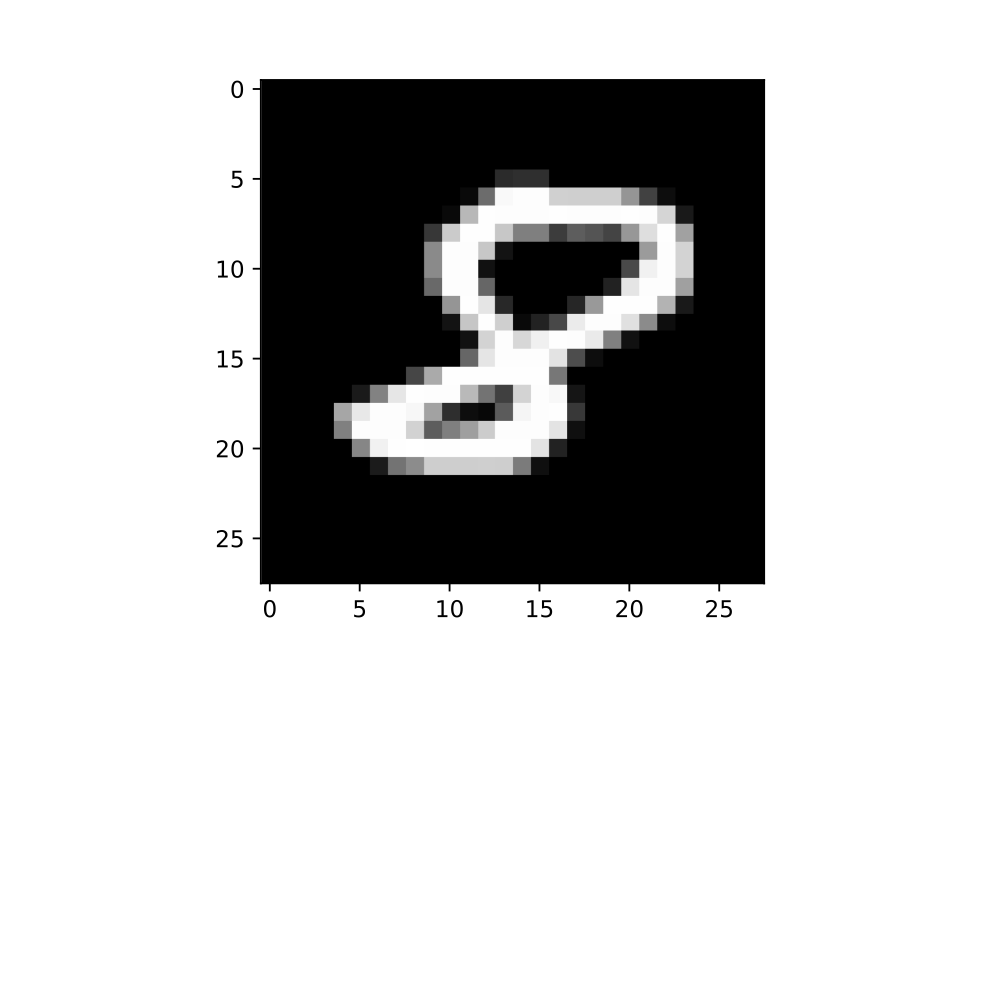}
  \includegraphics[width=.09\linewidth]{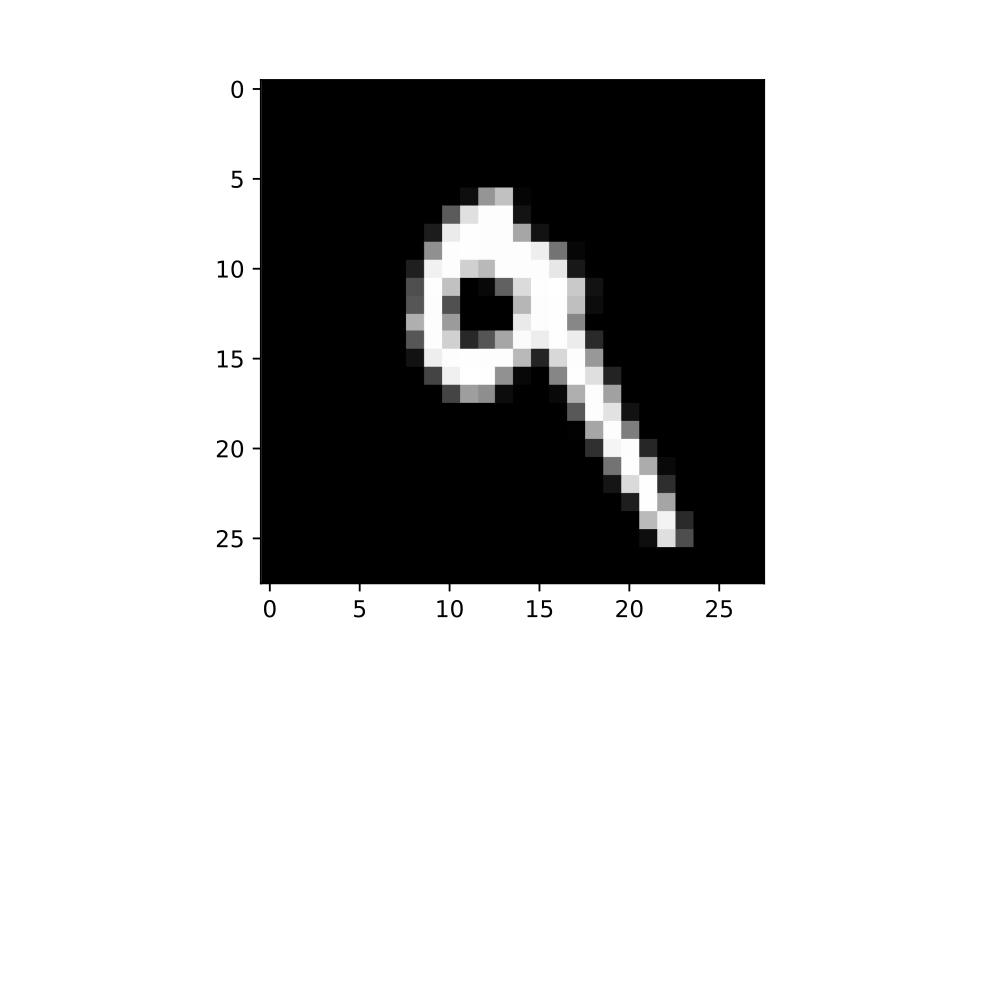}
\end{subfigure}
\begin{subfigure}{\textwidth}
  \includegraphics[width=.09\linewidth]{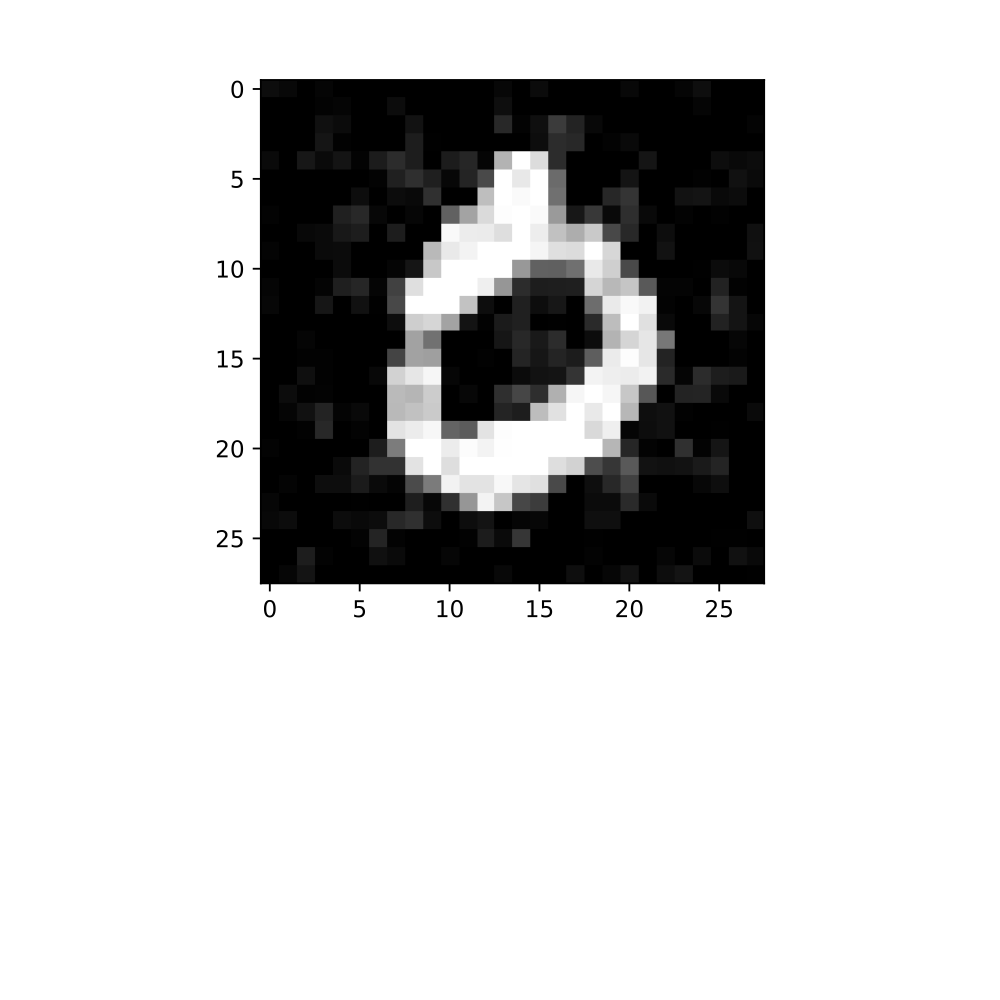}
  \includegraphics[width=.09\linewidth]{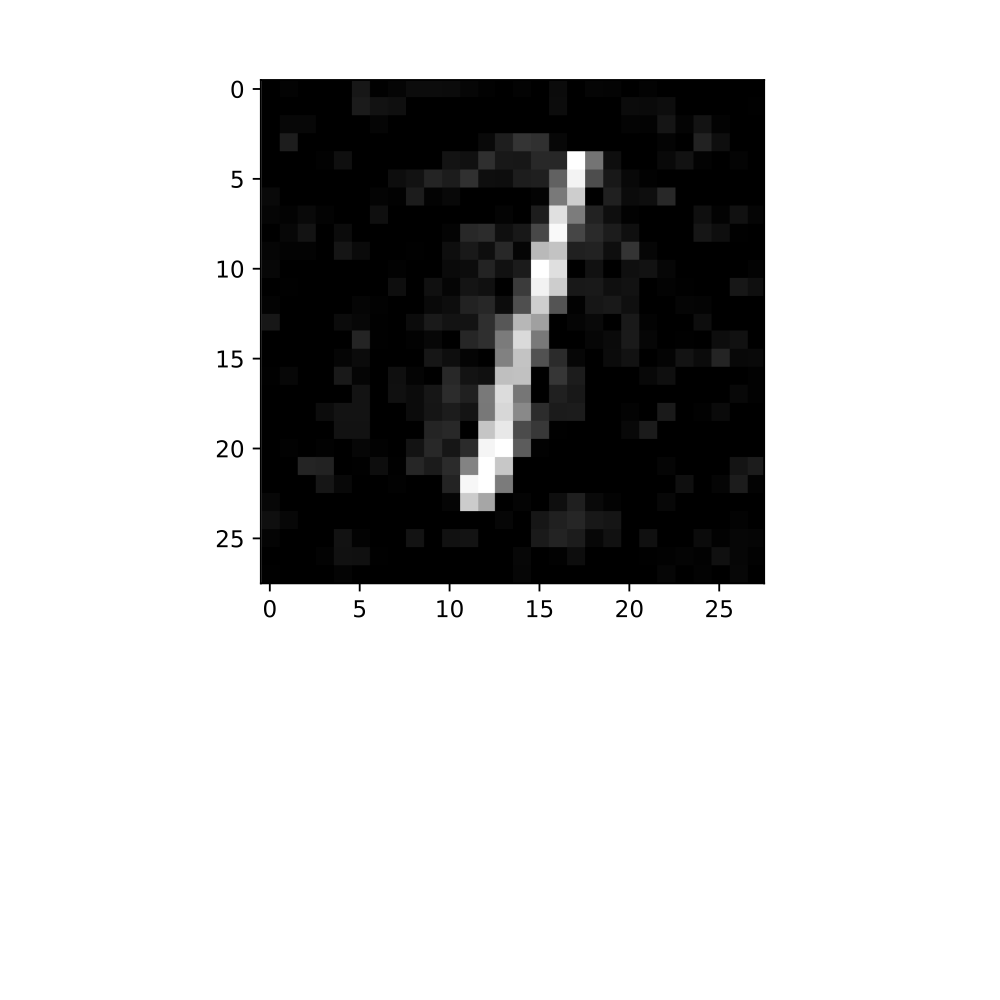}
  \includegraphics[width=.09\linewidth]{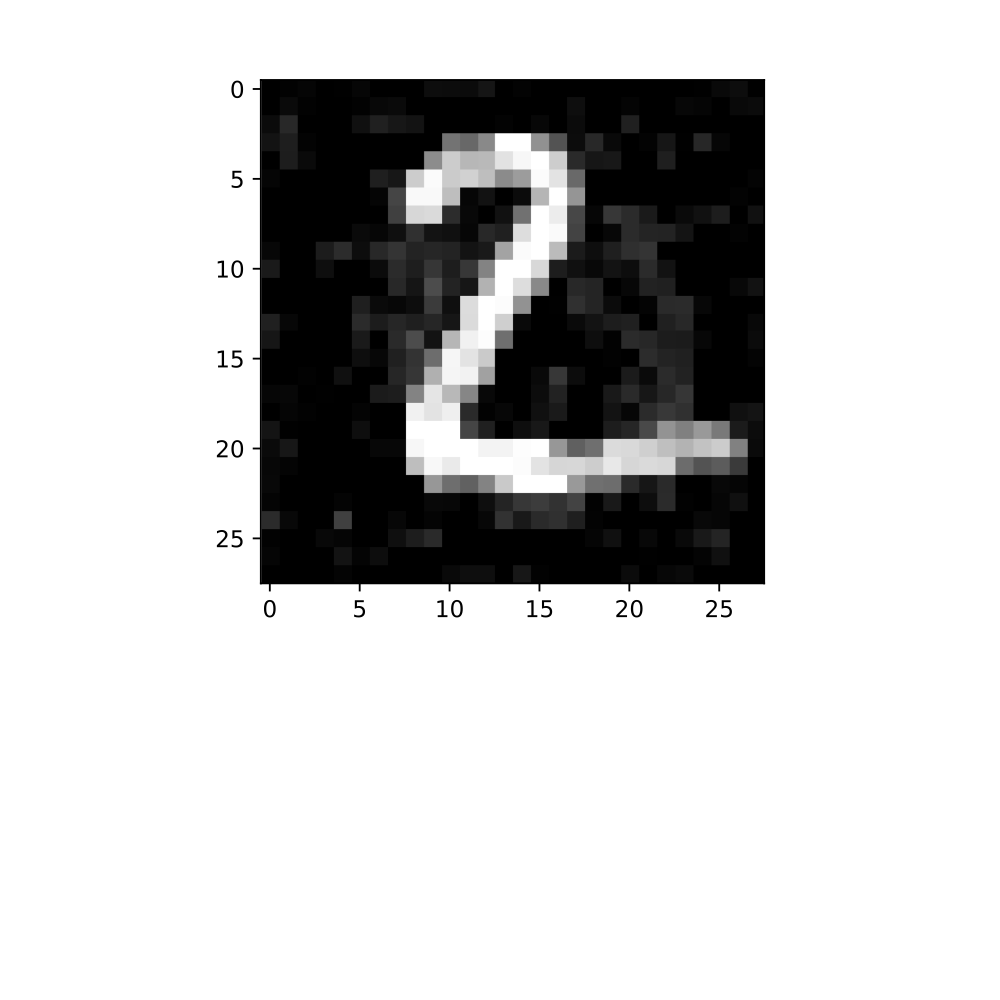}
  \includegraphics[width=.09\linewidth]{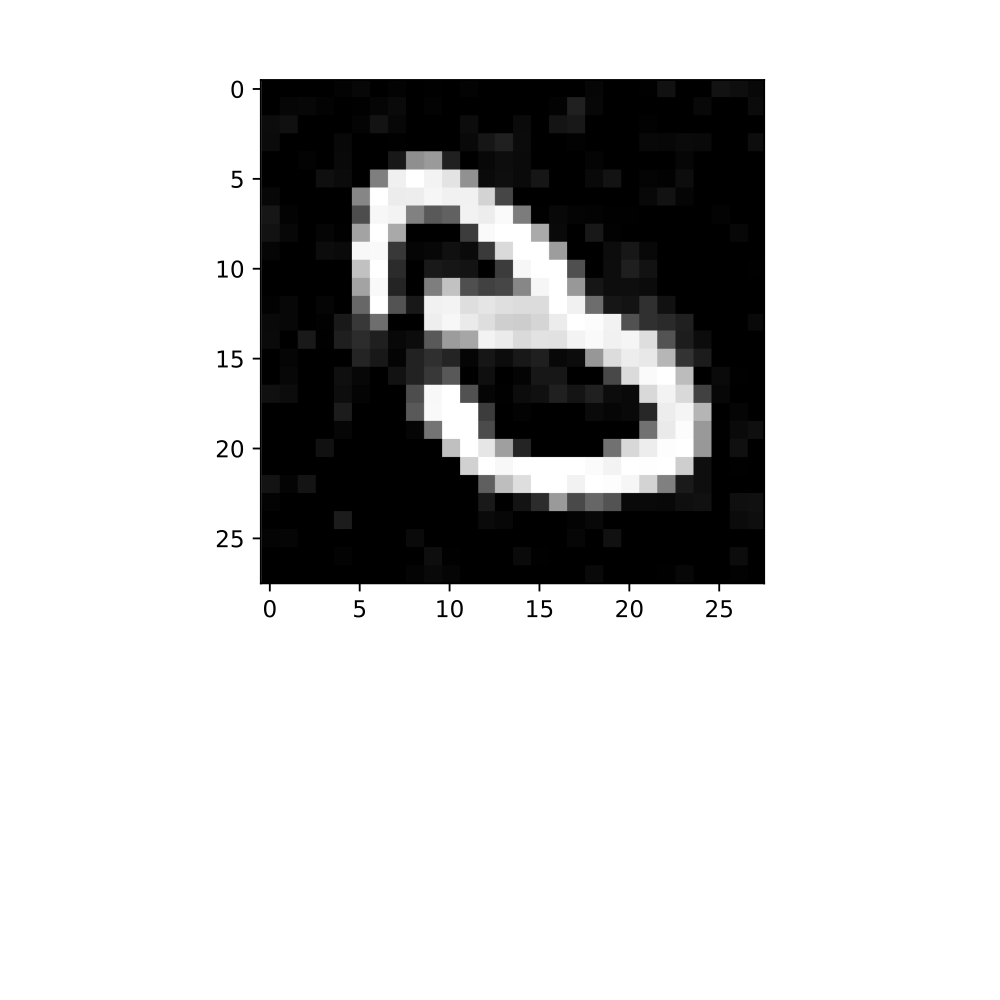}
  \includegraphics[width=.09\linewidth]{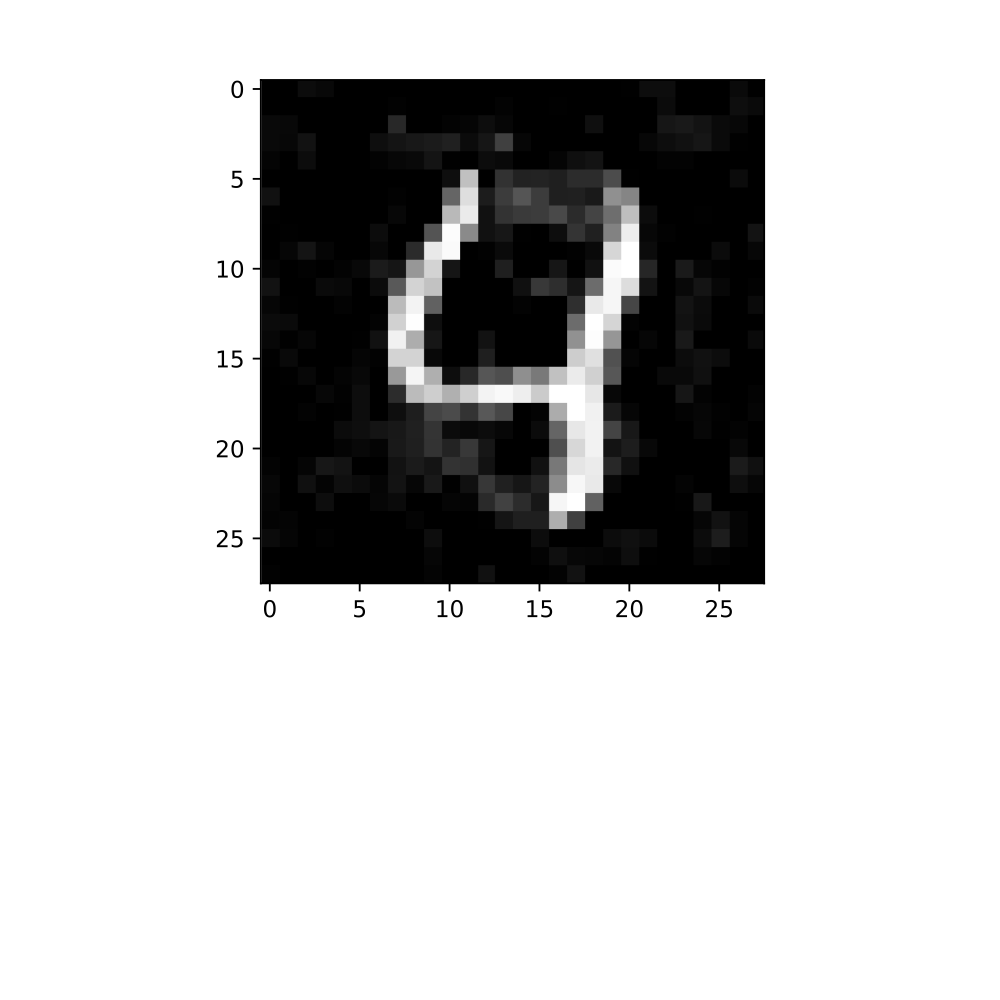}
  \includegraphics[width=.09\linewidth]{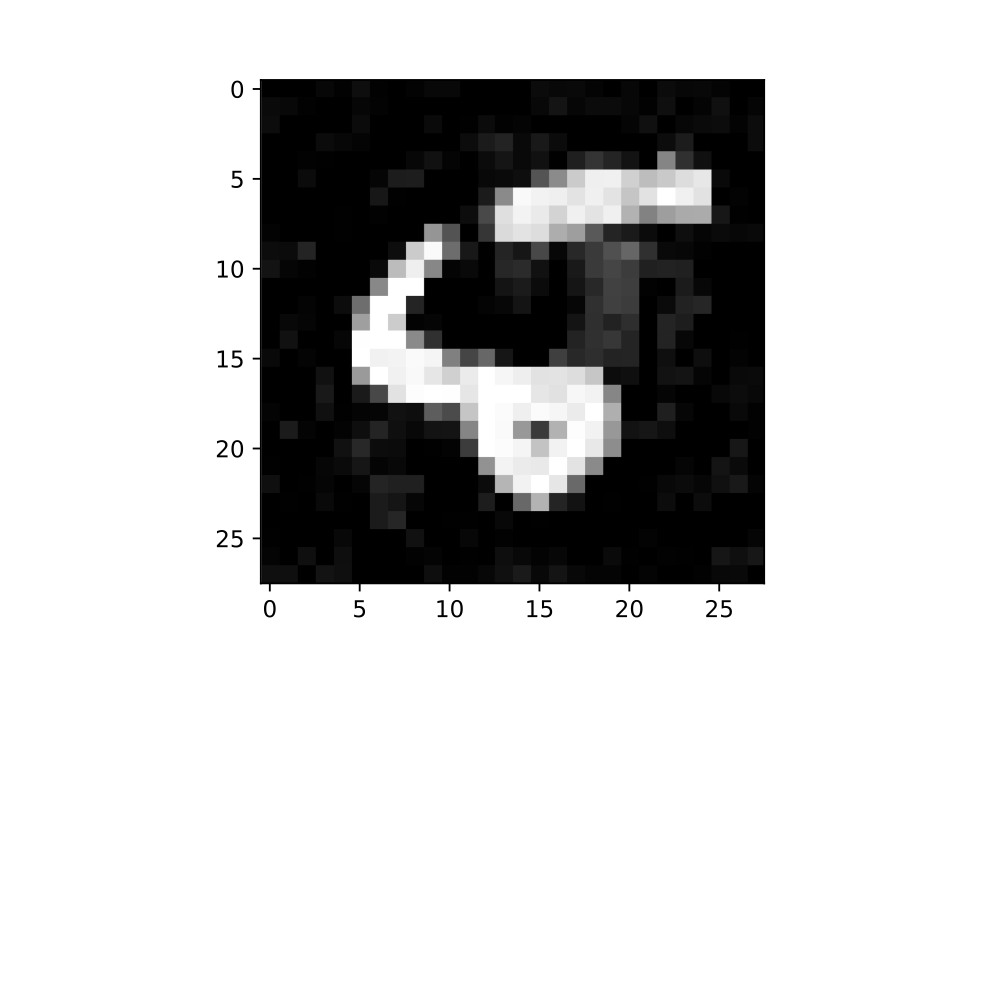}
  \includegraphics[width=.09\linewidth]{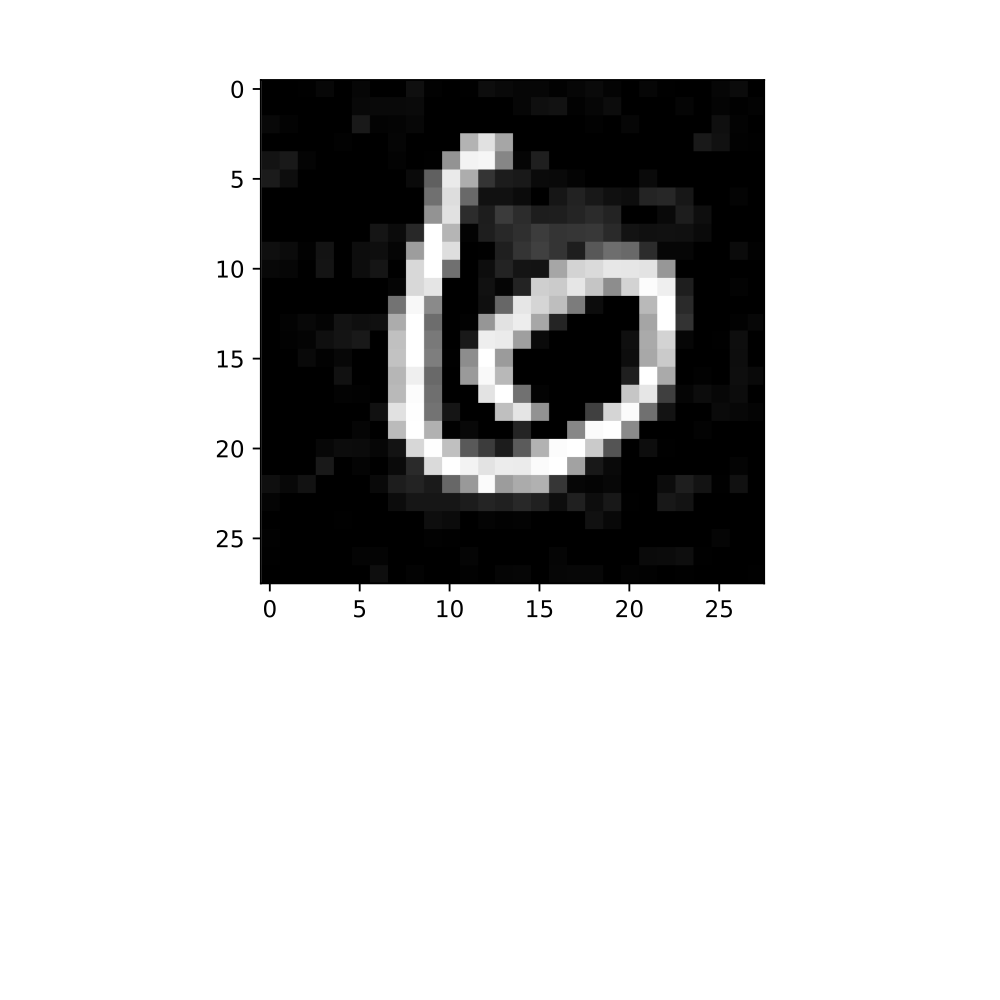}
  \includegraphics[width=.09\linewidth]{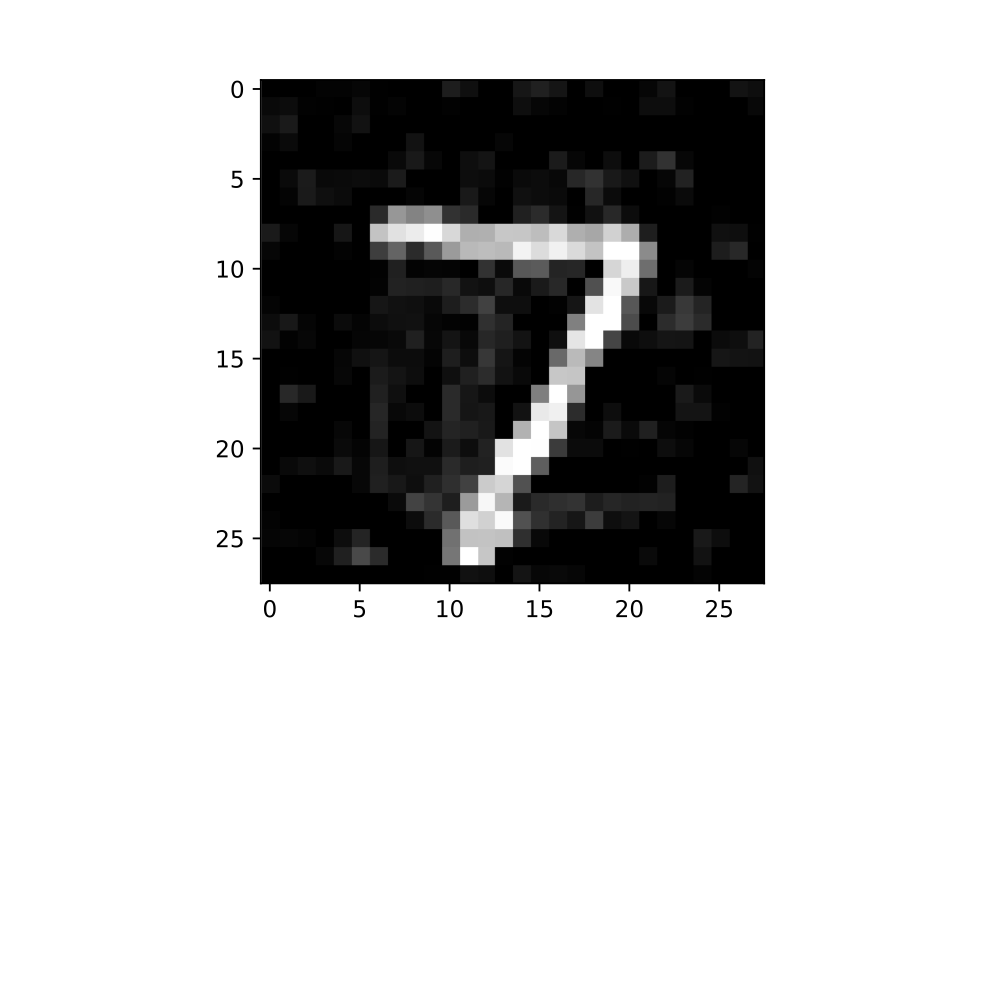}
  \includegraphics[width=.09\linewidth]{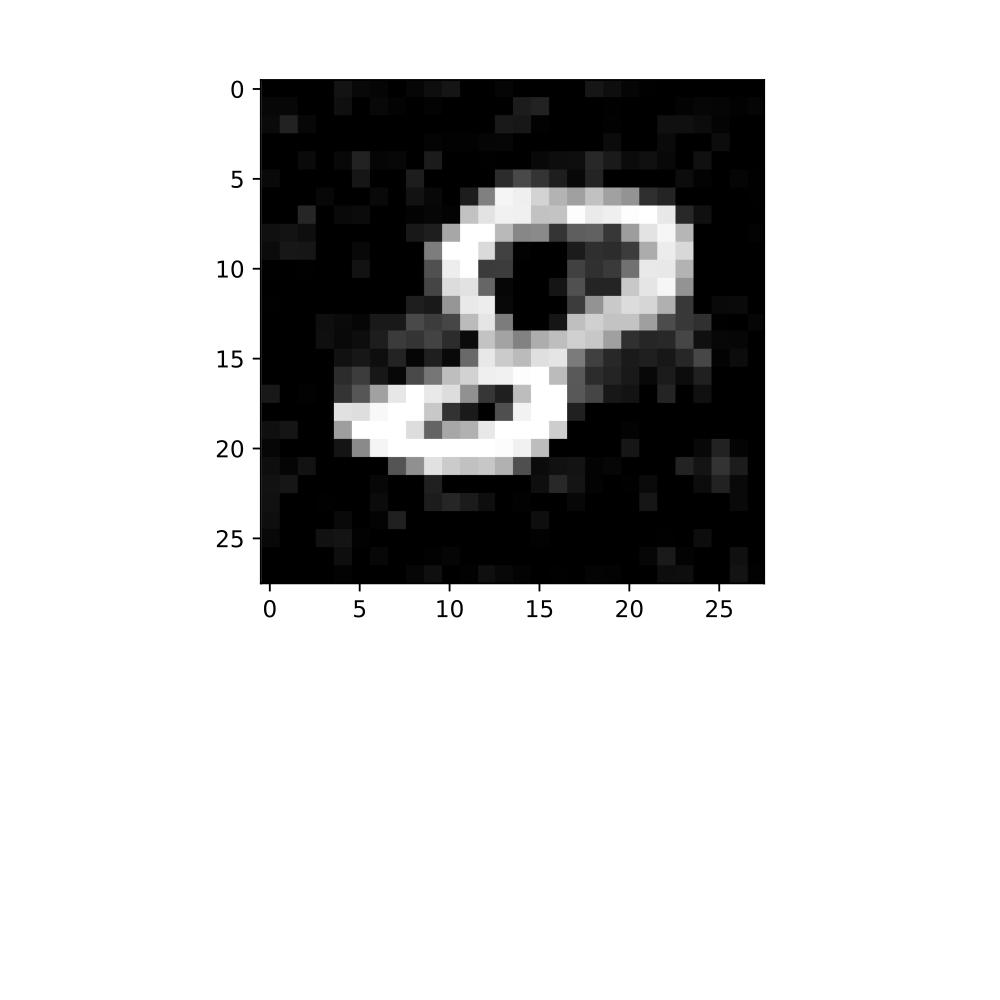}
  \includegraphics[width=.09\linewidth]{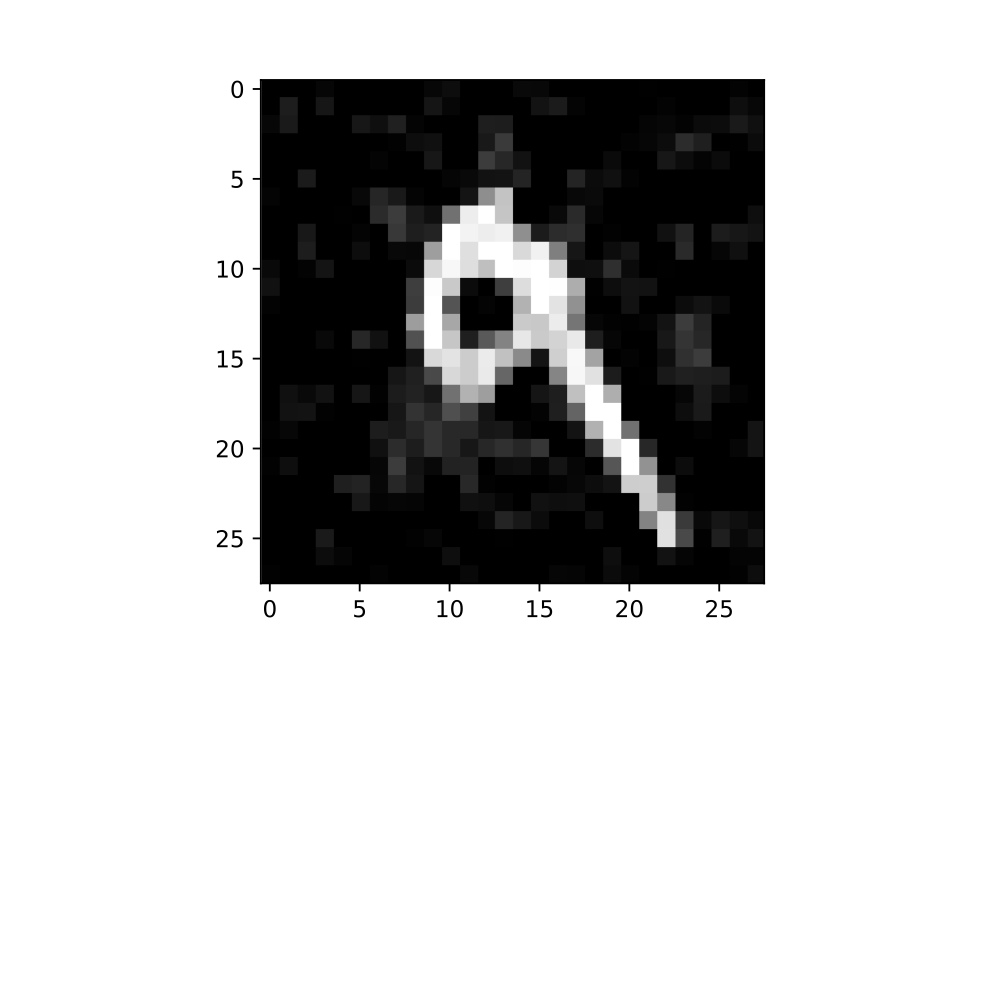}
 \end{subfigure}
\caption{\textit{Top}: Correctly classified, unaltered MNIST images. \textit{Bottom}: Adversarial images based off of the image directly above. Each adversarial image is misclassified as one digit higher modulo 9.}
\label{fig:mnists}
\end{figure}

We define an adversarial example as an input $\bm{x}'$ which is close to another input $\bm{x}$ according to some distance metric that, when input into a trained neural network, produces class prediction $C(\bm{x}') = t \not= C^*(\bm{x})$. That is, although $\bm{x}$ and $\bm{x}'$ are close in input space, $\bm{x}'$ causes the neural network to misclassify the input. $\bm{x}'$ in this scenario is referred to as a \textit{targeted adversarial example} with target $t$. 

The distance metric we focus on in this paper is the $L_2$ norm:

\begin{equation}
\| \bm{x'} - \bm{x} \|_2 = \sqrt{\sum\limits_{i=1}^n |x_i|^2}
\end{equation}

which measures the standard Euclidean distance between two vectors. There are other distance metrics that could be used ($L_0$ or $L_\infty$ norms, for instance). Analyzing performance differences in the detection algorithm described in this paper due to different distance metrics in adversary generation could be a route for future work, but as will become evident shortly, noticeable performance differences seem unlikely. 

The problem of finding an adversarial image is then characterized as:

\begin{align*}
\text{minimize} \ \ &\| \bm{x} - \bm{x} + \bm{\delta} \|_2 \\
\text{such that} \ \ &C(\bm{x} + \bm{\delta}) = t, \\
&\bm{x} + \bm{\delta} \in [0,1]^n
\end{align*}

In other words, we are looking to find a minimum distortion $\bm{\delta}$ that causes the network to misclassify the input $\bm{x}$ as class $t$. Following \cite{carlini2017towards}, we can solve this by choosing $\bm{w}$ that solves

\begin{equation*}
\text{minimize} \ \ \|\frac{1}{2}(\tanh(\bm{w}) + 1) - \bm{x} \|_2^2 + c f(\frac{1}{2}(\tanh(\bm{w}) + 1)
\end{equation*}

where

\begin{equation}\label{eq:minkappa}
f(\bm{x}') = \max(\max\{Z(\bm{x}')_i \ | \ i \not= t \} - Z(\bm{x}')_t - \kappa)
\end{equation}

The $\kappa$ parameter allows control of the confidence with which the neural network misclassifies the adversarial image. The $\bm{w}$ is optimized using gradient descent with multiple starting point images to better avoid local minima.

\section{Adversary Detection via Persistent Homology}

We describe in this section our method of adversary detection using only the topological signature of neural network computations induced by inputs. To gain intuition on the detection algorithm, we first provide some observations on the difference between topological signatures induced by unperturbed images versus adversarial images. 

\subsection{A Global Network Summary}

\begin{figure}
\begin{subfigure}{.33\textwidth}
\centering
  \includegraphics[width=.35\linewidth]{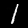} \vspace{5mm} \\
  \includegraphics[width=.75\linewidth]{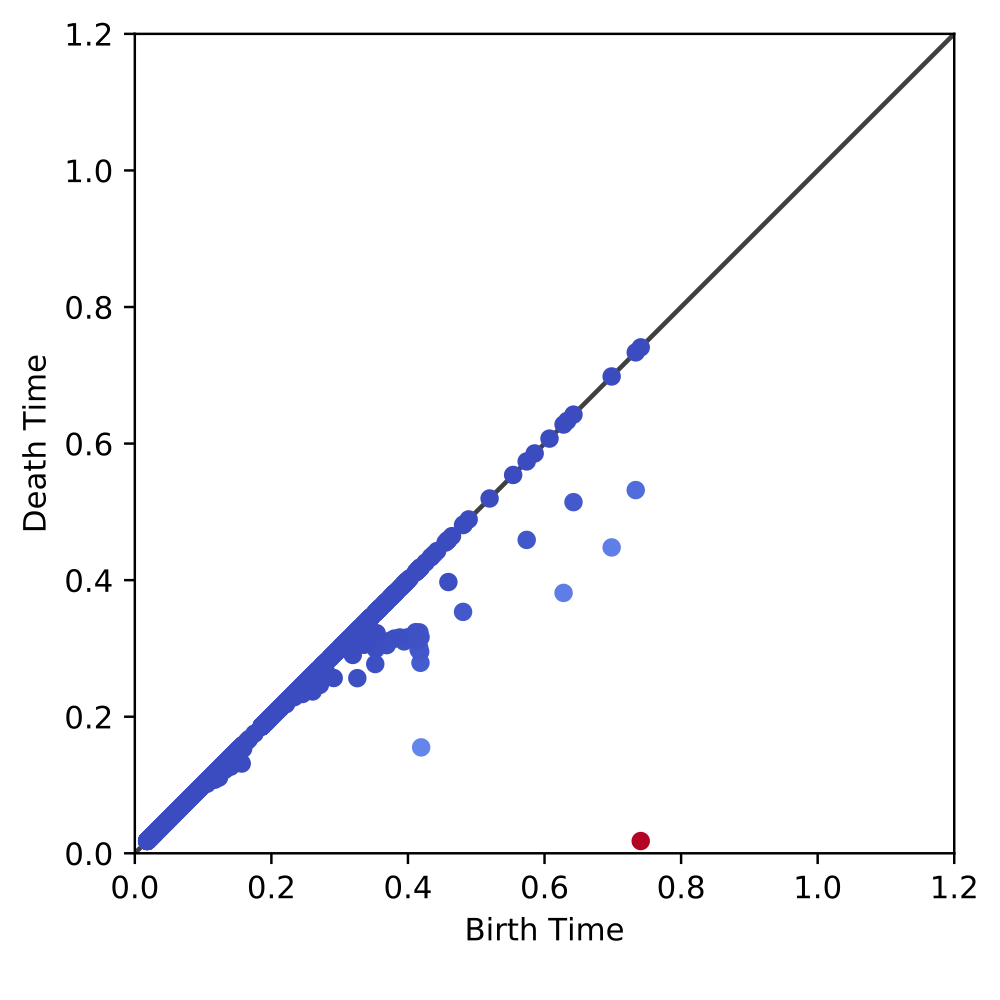}
\end{subfigure}
\begin{subfigure}{.33\textwidth}
\centering
\includegraphics[width=.35\linewidth]{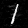} \vspace{5mm} \\
 \includegraphics[width=.75\linewidth]{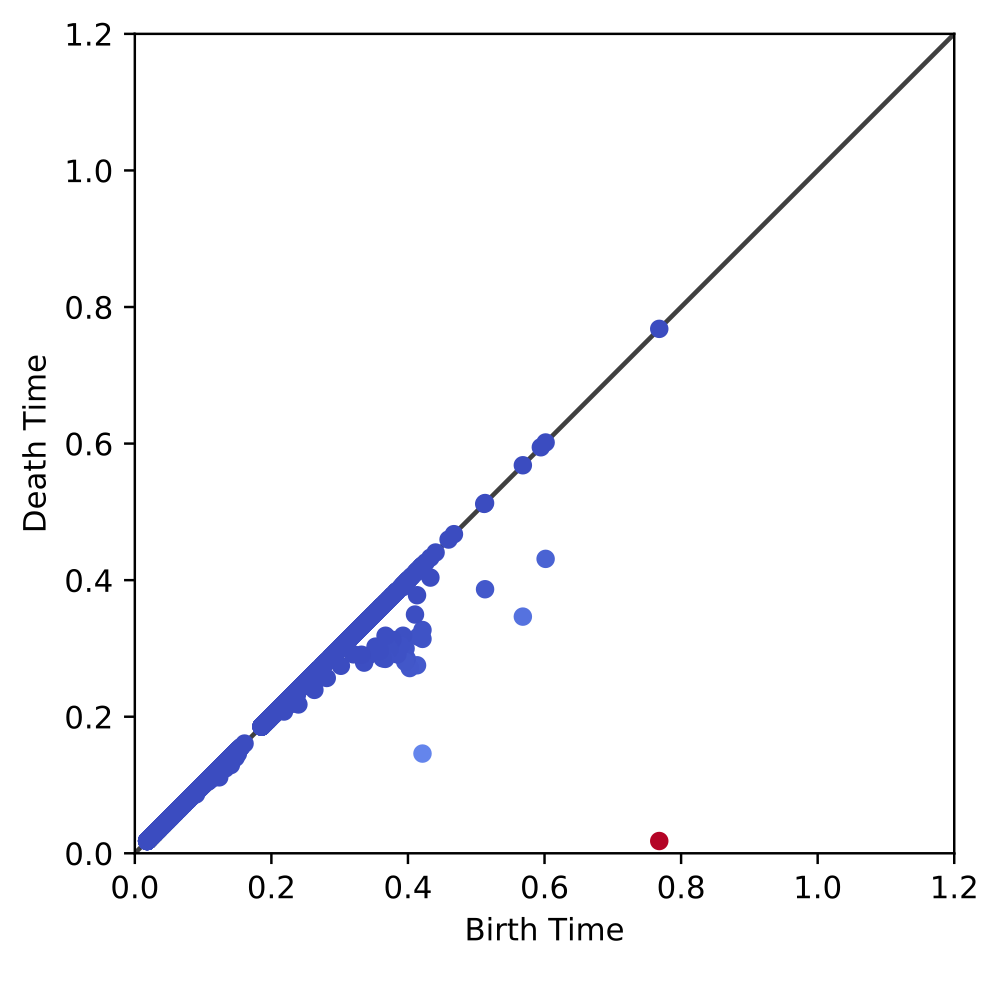}
 \end{subfigure}
\begin{subfigure}{.33\textwidth}
\centering
  \includegraphics[width=.35\linewidth]{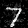} \vspace{5mm} \\
  \includegraphics[width=.75\linewidth]{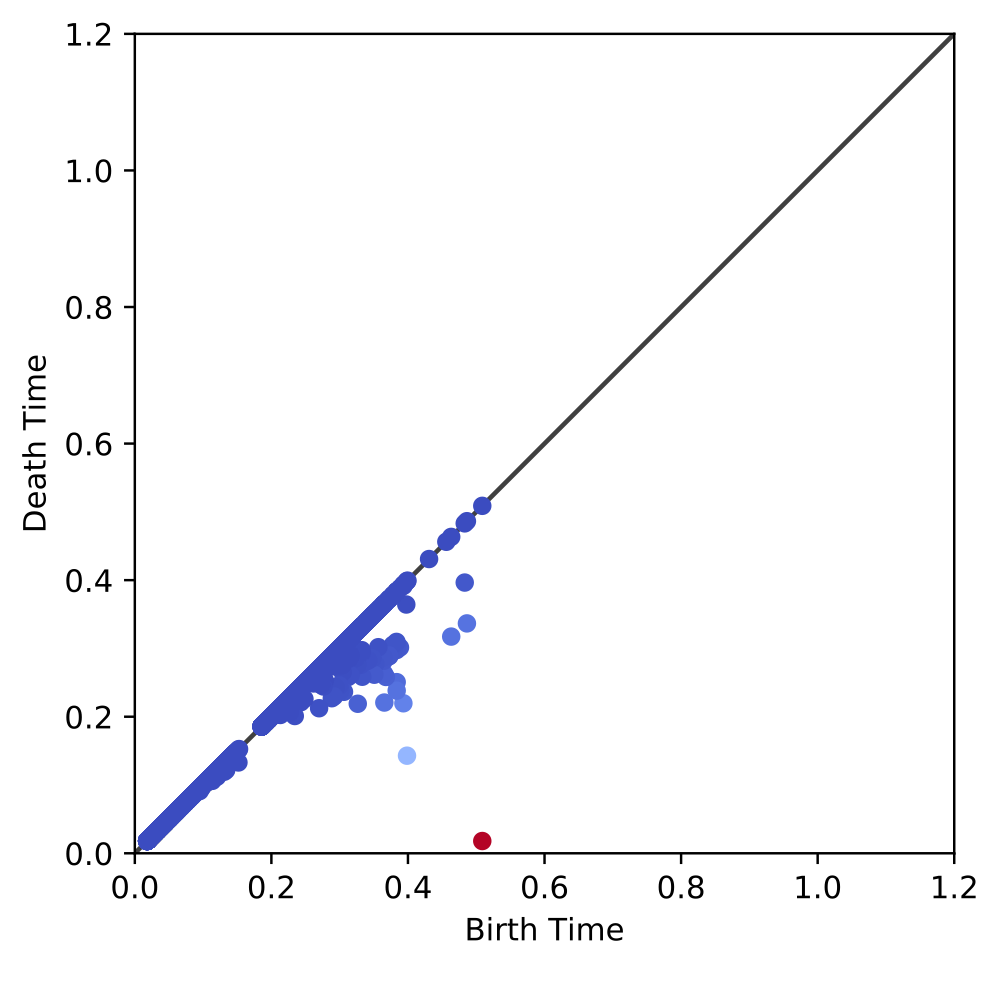}
 \end{subfigure}
\caption{Persistence diagrams induced by similar images are similar. Despite the first two images being classified by the neural network as different classes, their persistence diagrams appear more similar than the third image which is classified the same as the first image. \textit{Left}: The original, correctly-classified MNIST image and its corresponding persistence diagram. \textit{Middle}: An adversarial image generated from class 1 that is misclassified as a 7 by the neural network and the corresponding persistence diagram. \textit{Right}: An adversarial image generated from class 7 that is misclassified as a 1 and its corresponding persistence diagram.}
\label{fig:interpolations}
\end{figure}

Persistent homology provides a robust summary of the connected components of a neural network computation across varying levels of granularity. This summary can be represented as a persistence diagram (Figure \ref{fig:interpolations}). In our usage, each point in the persistence diagram corresponds to a subgraph within the neural network graph that exists as a a disconnected component between the time $[b,d]$ where $b$ is the edge threshold value corresponding to the birth time and the death time $d$ which corresponds to the edge weight threshold at which the component joins a higher-dimensional component. The component with birth $b$ and death $d$ has lifetime $b - d$.

With this understanding, it is natural to ask whether the topological structure (and therefore the persistence diagrams) of neural network computations induced by adversarial examples differs from those induced by unperturbed inputs. Figure \ref{fig:interpolations} shows there are indeed differences in persistence diagrams induced by different inputs. In fact, we find that the Wasserstein distance between persistence diagrams tracks nearly continuously as we interpolate linearly between examples in input space. 

Unfortunately, persistence diagrams are not unique in their representation of the underlying network that generates them. Because of this, two different networks induced by two different inputs can end up having similar persistence diagrams even though the points in each diagram are generated by completely different subgraphs within the neural network. However, the subgraphs generating these points are uniquely identified with respect to the neural network itself. Therefore, two persistence diagrams that have the same points in the $(b,d)$ plane which are generated by the same subgraphs in the network would  be equivalent. Thus, we look to detect adversarial inputs based on the most robust (``most persistent") subgraphs that exist in the input-induced networks. In other words, we are looking to detect adversaries by noting which semantic signatures of the input-induced network do not match the classification made by the neural network. We look to discriminate substructures within the neural network graph that correspond to adversarial examples versus unperturbed examples. 

This approach aligns with the current understanding of adversarial examples and the representations built by neural networks. It is generally accepted that convolutional neural networks make classification decisions by building up increasingly complex object representations across layers \cite{zeiler2014visualizing}. The semantic information of these representations are represented distributively across many neurons \cite{li2015convergent}, and it is understood that a neuron may activate in numerous different semantic scenarios \cite{}. Our adversary detection algorithm relies on these properties of neural networks in that the topological signature of an input-induced network is constructed from the ensemble of the most robust substructures (neurons and connections) across all thresholds of activation values.  For two images that are semantically similar (qualitatively or in terms of total distortion), their topological signatures should be similar if the network is making use of robust semantic representations in its classification decisions.

In a recent paper, Cubak et al. \cite{cubuk2017intriguing} observe that adversarial examples may be artifacts of the lack of variance across logit values. The paper investigates adversaries produced by the Fast Gradient Sign Method (FGSM), but the same conclusions may be drawn from the adversarial generation process used in this paper. Namely, these adversary generation algorithms target the logit values output by the neural network and, subsequently, the network's inherent uncertainty in its classification decision. We believe our adversary detection method is robust to adversary generation algorithms (and their choice of loss), as our algorithm uses information derived from the computations throughout the entire network, not just the information in the output or pre-softmax layers. Our detection algorithm is dependent on the substructures used to classify unperturbed images being noticeably different than those used to classify adversarial images. Therefore, we expect this algorithm to perform well in networks that have built robust representations for use in classification decisions but poorly in undertrained or networks that have memorized training data as this robust semantic information will not be picked up in these cases. Similar reasoning leads us to expect our detection algorithm to work \textit{better} as total distortion increases for adversarial examples.

\subsection{Model}

We train a simple convolutional neural network on the MNIST handwritten digits dataset. Our network consists of a single convolutional layer followed by a $25088 \times 1024$ fully-connected layer and a $1024 \times 10$ fully-connected layer. The convolutional layer contains 32 $5 \times 5$ filters each with stride 1 and same padding. The weight variables on the fully-connected layers are initialized with truncated normal with standard deviation $0.1$. The bias variables are initialized similarly.  During training, we randomly dropout half the connections according to a uniform normal distribution. Our model achieves 98.7\% accuracy on MNIST after 50 epochs of batch size 128.  

\subsection{Persistent Homology Calculation}

Using notation from Section \ref{sec:adversaries}, our trained neural network model $F: \mathbb{R}^n \rightarrow \mathbb{R}^m$ takes an input example $\bm{x} \in \mathbb{R}^n$ and maps it to a probability distribution over classes $\bm{y} \in \mathbb{R}^m$. However, as we saw in Section \ref{sec:networkgraph}, this network may also be fully represented as a weighted, directed graph. Denote this graph $G = (V,E,w)$ where $V$ are the vertices (neurons), $E$ the edges (layer-wise connections), and $w: E \rightarrow \mathbb{R}$ a mapping from edge to edge weight (layer connection to weighted connection). In other words, the edges are represented by neural connections between layers and their weights are derived from the connection weights that are described by, for example, the weight matrices in fully-connected layers or the filters in convolutional layers. For a given architecture, $V$ and $E$ are fixed. During training, $w$ changes as the network learns to represent features of the input space. For the rest of the paper, we assume the neural network has been trained such that $w$ remains fixed. 

For a given input $\bm{x}$, feeding this input into the neural network induces a new graph $G_{\bm{x}} = (V_{\bm{x}}, E_{\bm{x}}, f)$. $G_{\bm{x}}$ is a subgraph of $G$ in that its edges and vertices are derived from $G$ such that $V_{\bm{x}} \subseteq V$ and $E_{\bm{x}} \subseteq E$. The function $f : E_{\bm{x}} \rightarrow \mathbb{R}$ is a new mapping on the edge weights such that $f = w \circ g$ for some relationship between input and edges $g : \mathbb{R}^n \rightarrow E$ that depends on the network and input. Here, we assume that edges that have zero weight (i.e. $f(e) = 0 \ \text{for} \ e \in E_{\bm{x}}$ are not present in $G_{\bm{x}}$. Thus, depending on the activation functions used, $V_{\bm{x}}$ and $E_{\bm{x}}$ may be proper subsets of $V$ and $E$. For example, ReLU activation would lead to numerous edges in $G$ being sent to 0 in the induced subgraph $G_{\bm{x}}$.

We can view $G_{\bm{x}}$ as a geometric realization of a finite-dimensional simplicial complex $K$ if we assume an embedding of the graph into Euclidean space. We would like to compute the persistent homology of this simplicial complex. However, our current input-induced graph can contain edges with weight less than zero. While this does not impede the persistent homology calculation, it does affect our intuition of how edge weights correspond to information flow through our input-induced graph. Namely, edges with large negative weights are passing a large-magnitude suppressant signal into the next layer. We would like to pick up on this information. As we will see, the persistent homology calculation requires a filtration on the vertices and edges of its simplicial complex, and most natural filtrations on a graph with negative and positive edge weights would lose the semantic importance of large negative edge weights. Therefore, we define a new graph $G^*_{\bm{x}} = (V_{\bm{x}}, E_{\bm{x}}, \phi)$ where $\phi = | f |$ by taking the absolute value of edge weights in $G_{\bm{x}}$. The absolute value edge weights lead to an equal semantic interpretation between large positive activation information and large negative activation information as both are, in theory, more important to classification decisions made by the network than small negative or positive activation information. We can view $G^*_{\bm{x}}$ as a geometric realization of a simplicial complex $K^*$.

We now look to compute the most persistent features in $G^*_{\bm{x}}$ by computing the persistent homology of $K^*$. We consider the construction of a Rips complex on $K^*$. For $r \in \mathbb{R} \ \text{and} \ r > 0$, a Rips complex $R(r)$ is formed by considering the ball of radius $\frac{r}{2}$ centered at vertices in $K^*$. A 1-simplex (edge) is formed between two vertices in $K^*$ if and only if their balls intersect. A 2-simplex (triangle) is formed among three points if and only if all three balls intersect. 

\begin{wrapfigure}{r}{0.35\linewidth}
\centering
  \includegraphics[width=\linewidth]{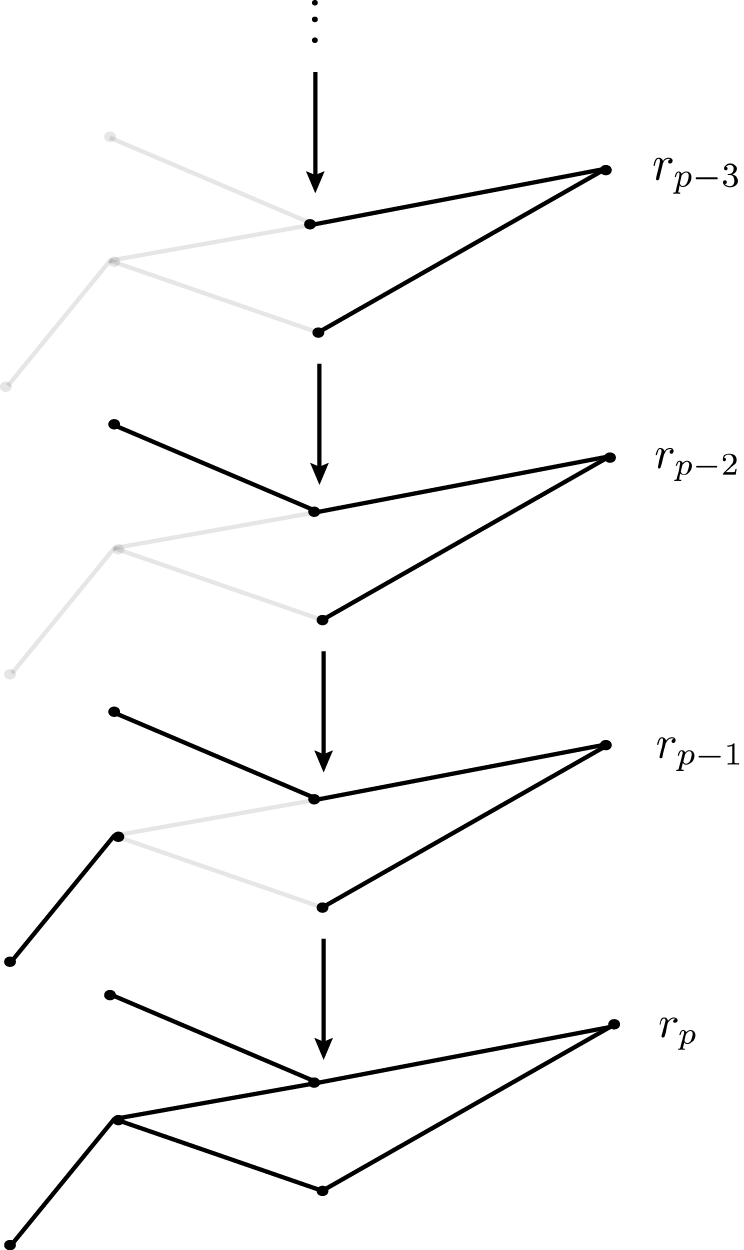}
\caption{Simple example of network filtration}
\label{fig:filtration_demo}
\end{wrapfigure}

Let $\omega = \text{max}(\phi)$ be the maximum edge weight in the graph. We embed $G_{\bm{x}}$ in Euclidean space by assuming the pairwise distance between each node in $V_{\bm{x}}$ to be represented by the difference between $\omega$ and its edge weight. In other words, nodes connected to edges with high weight are closer than nodes connected by a low-weight edge. We can view this embedding as a new graph $G^*_{\text{emb}} = (V_{\bm{x}}, E_{\bm{x}}, \phi')$ with the same vertex and edge sets but with $\phi' = \omega - \phi$. If we consider a finite increasing sequence $0 = r_0 \leq r_1 \leq r_2 \dots \leq r_p$, then the Rips filtration $\mathscr{F}$ is a sequence of Rips complexes connected by well-defined inclusions $R(r_0) \xhookrightarrow{} R(r_1) \xhookrightarrow{} R(r_2) \xhookrightarrow{}{} \dots \xhookrightarrow{} R(r_p)$. The inclusions induce homomorphisms between the homology groups of $\mathscr{F}$ such that $H_*(R(r_0)) \rightarrow H_*(R(r_1)) \rightarrow H_*(R(r_2)) \rightarrow \dots \rightarrow H_*(R(r_p))$.

We are only interested in $H_0$ and $H_1$ in this scenario as a feedforward networks only produce--at largest--2-dimensional objects (diamonds or other polygons) from their edge connections, leading to trivial $H_i$ for $i > 1$. $H_0$ tracks 1-dimensional topological components which correspond to connected components within our neural network.

Topological features like connected components or polygons appear and disappear as $r$ increases. The value $b = \omega - r$ at which a component appears is its \textit{birth} time. The value $d = \omega - r$ at which a topological feature disappears (a connected component merges with another or a polygonal hole is filled) is known as the component's \textit{death} time. We can then define the lifetime of a component as $l = b - d$. Again, we take $\omega - r$ because we hold the intuition that components with higher edge weights are more semantically meaningful in classification decisions made by the neural network than components with lower-weight edges. 

Because $H_2$ is trivial, most $H_1$ components of our input-induced network do not get filled into higher-dimensional simplices and thus have an infinite lifetime. However, we elect to truncate these lifetimes at the minimum over all edge weights $\text{min}(\phi)$ as this is technically the end of the filtration process. This also makes analysis easier as these are endowed with lifetime $l = b - \text{min}(\phi)$ instead of $l = b - (-\infty)$ as would otherwise be the case. 

\begin{wrapfigure}{r}{0.35\linewidth}
\centering
  \includegraphics[width=\linewidth]{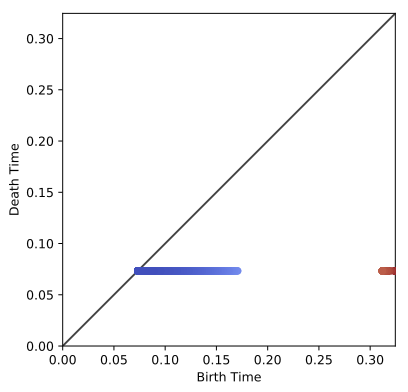}
\caption{Persistence diagram of $H_1$}
\label{fig:H_1}
\end{wrapfigure}

We found in our analysis that information from $H_1$ is not especially meaningful for adversary detection. We believe this is due to the fact that almost all $H_1$ components are infinitely lived, and are thus truncated to $d = \text{min}(\phi)$. This can be seen in Figure \ref{fig:H_1}. Because there are few 2-dimensional components being integrated with other 2-dimensional components, there is not much information to be gained on which components are more robust than others. As well, it is unclear to the authors what the semantic meaning of a persistent 2-dimensional component of a feedforward neural network would be. Because of this, we focus on 1-dimensional components (connected components) that are tracked by $H_0$. As a note, if we were to apply this method to a recurrent architecture, then the $H_1$ homology group would track, at least in part, the recurrent relationships within the network. This is an interesting area for future work.

The output of the above persistent homology calculation on $G_{\bm{x}}$ is a set $D \subset \mathbb{R}^2$ of birth-death pairs $(b,d) \in D$ which may be represented as a persistence diagram in the $(b, d)$ plane. See Figure \ref{fig:filtration_demo} for a visual representation. Equivalently, this persistent homology computation is a map\footnote{This map is bijective in practice. Our construction of $D$ leaves room for two pairs $(b_1, d_1)$ and $(b_2, d_2)$ such that $b_1 = b_2$ and $d_1 = d_2$. In this case, computing persistent homology would give us two points $(b,d)_1$ and $(b,d)_2$ that are unique even though their birth and death components are equivalent. We can easily map this back to a unique subgraph in practice.} $\tau: G_{\bm{x}} \rightarrow D$ from subgraphs to points in $D$. Each of these points is associated to a connected component in $G_{\bm{x}}$ between edge weight resolutions $b$ and $d$. In other words, we can compute an inverse $\tau^{-1}: D \rightarrow G_{\bm{x}}$. 

Intuitively, the components with the highest lifetime $l$ should be those most strongly associated to classification decisions, representing substructures within the neural network that are activated based on prominent features within the input. We hypothesize that differences in these components should be observable for adversarial inputs when compared to non-adversarial inputs of the same predicted class.

Take $\lambda$ to be the threshold on the lifetime of points in the persistence diagram. We then calculate the set of weighted subgraphs $G^\lambda_{\bm{x}}$ of $G_{\bm{x}}$ that correspond to each $(b,d) \in D$ such that $b - d > \lambda$ such that $G^\lambda_{\bm{x}} = \{G_l \ | \ G_l = \tau((b,d)), \ b - d = l > \lambda \}$. This set $G^\lambda_{\bm{x}}$ is actually a subgraph of $G_{\bm{x}}$. It is this subgraph $G^\lambda_{\bm{x}} = (V^\lambda_{\bm{x}}, E^\lambda_{\bm{x}}, \phi')$ that will serve as the basis for detecting adversarial inputs in neural networks.

\subsection{Adversary Detection}

We describe in this section three related adversary detection algorithms based on the persistent subgraphs obtained from input-induced neural network graphs $G_{\bm{x}}$. 

To begin, note that most neural networks contain upwards of millions of nodes and edges. Even with ReLU activations dropping out edges, computing persistent homology on such large graphs is computationally burdensome. Thus, we introduce parameter $\rho \in [0,1]$ which acts as the layer-wise percentile threshold on edge weights prior to the persistent homology computation. In other words, we take only the top $100\rho$\% largest edge weights for \textit{each layer}. We perform this calculation layer-wise because average edge weight may be vastly different between layers. In fact, we see this being the case such that, in general, convolutional layers have higher edge weights than fully-connected layers. We set $\rho = 0.99$ for the results described below. One could argue that only taking the top 1\% of weighted edges throws away a lot of information. As we shall see, a lot of the relevant information is retained in the top 1\% of edges which implies many of the induced edge weights may be noise.

We have two parameters, $\lambda$ and $\rho$, to control the complexity of the persistent subgraph $G^\lambda_{\bm{x}}$ induced by an input. The $\lambda$ parameter controls the minimum lifetime of the components in $G^\lambda_{\bm{x}}$ while $\rho$ controls the weight distribution and number of edges that are used in the persistent homology calculation.

For our experiments, we construct two sets of adversarial images from the MNIST dataset using the method described in Section \ref{sec:adversaries} and code provided in \cite{carlini2017towards}. For each set, we sample 50 images from the MNIST dataset and construct 9 targeted adversaries from each for a total of 450 adversarial examples. The first set of 450 adversarial images, $A_{0}$, are produced with $\kappa = 0$, resulting in average distortion of 1.83. The second set, $A_{20}$, was produced with $\kappa = 20$, leading to high-confidence predictions in the target class for the adversarial examples and average distortion 3.54. We also sampled 1000 unaltered images from the original dataset. We denote this set of images as $U$. Of these images, 550 were used for training and 450 for testing. We denote these $U_{train}$ and $U_{test}$, respectively such that $U_{train}, U_{test} \subset U$. We assume we have access to the correct class labels for $U_{train}$ such that we know $\text{class}(\bm{x})$ for $\bm{x} \in U_{train}$.

\subsubsection{Maximum Node Matching} \label{sec:max_node_matching}

In this detection method, we look to detect discrepancies between the input-induced topological signature and the expected topological signature of the predicted class. In essence, we are looking for the semantic substructures within the network that correspond to the true class of the input despite the fact that the adversary may have added a sufficiently small amount of distortion of the image to cause the predicted class to change. 

To do this, we fix $\rho$ and $\lambda$ and compute the persistent subgraph $G^\lambda_{\bm{x}}$ for each $\bm{x} \in U_{train}$. For each class $y_i \in \bm{y}$, we aggregate all subgraphs computed for each training image with that class label. Thus, for each $y_i \in \bm{y}$ we have a set $\mathscr{G}_i$ of subgraphs of $G_{\bm{x}}$ where $\mathscr{G}_i = \{G^\lambda_{\bm{x}} \ | \ \bm{x} \in U_{train}, \text{class}(\bm{x}) = i\}$. Let $\mathscr{V}_i$ be the set of vertices of $\mathscr{G}_i$. Remember, $\mathscr{V}_i \subset V_{\bm{x}}$. We then count the occurrences of each unique vertex in $\mathscr{V}_i$ for each $i$ and rank the now unique vertices according to their appearance count. Thus, we have a function $r_i: \mathscr{V}_i \rightarrow \mathbb{Z}$ surjectively taking each $v \in \mathscr{V}_i$ to its occurrence rank. 

Now, for each class $i$ we have a ranked occurrence list for vertices of $V$ that appear in the induced persistent subgraphs $G^\lambda_{\bm{x}}$ for each $\bm{x} \in U_{train}$. We hypothesize that these vertex sets are an indicative aggregate of the most persistent semantic subgraphs of $G$ used for classification in each class. Thus, images of class $i$ should induce persistent subgraphs that are most similar to the aggregate vertex set $\mathscr{V}_i$. We define this similarity as the normalized sum in vertex occurrence ranks between $\mathscr{V}_i$ and the induced subgraph vertex set $V^\lambda_{\bm{x}}$. Therefore, we define our match function as

\begin{equation}
m_i(v) = 
\begin{cases} 
      r_i(v) & v \in \mathscr{V}_i \\
      0 & \text{otherwise} 
   \end{cases}
\end{equation}

For each class $i$, we then define a similarity score

\begin{equation}
	s_i(G^\lambda_{\bm{x}}) = \frac{\sum\limits_{v \in G^\lambda_{\bm{x}}}m_i(v)}{| \mathscr{V}_i | }
\end{equation}

We can then find the class of highest similarity via $i^* = \argmax_i(s_i(G^\lambda_{\bm{x}}))$. This $i^*$ is the class whose topological signature most resembles the topological signature of the persistent subgraph induced by $\bm{x}$.

Recall our network $F$ takes image $\bm{x}$ and returns a distribution over classes $\bm{y}$. We take the maximum probability over classes $\hat{i} = \argmax_i(\bm{y})$ as our class prediction. Thus, for each input $\bm{x}$, we view the network class prediction $\hat{i} = \argmax_i(F(\bm{x}))$.

Our adversary detection scheme is to simply note differences in $\hat{i}$ versus $i^*$. Specifically, if $\hat{i} \not= i^*$ we flag the input as a potential adversary. We test this adversary detection scheme on $A_0$ by iterating over $A_0$ and $U_{test}$, computing $i^*$ and $\hat{i}$ for each image. If $\hat{i} \not= i^*$, we mark the image as an adversary. The results of this process are shown in Table \ref{table:max_match}.

\subsubsection{Average Node Matching}

In defining the maximum node matching detection algorithm, we noticed that adversarial inputs generally have higher similarity scores to \textit{all} classes than non-adversarial inputs. This section describes an adversary detection algorithm motivated by this observation. 

We must first compute the expected matching of non-adversarial inputs to be able to compare this to potential adversarial inputs. We first remove a subset $U_{val}$ from the training set $U_{train}$ that will serve to calculate the expected similarity across all classes for the training set. To construct this expected match value, we follow the same matching construction as in Section \ref{sec:max_node_matching} except instead of taking the maximum matching class, we take the mean number of matches over all classes:

\begin{equation}
s_{avg}(G^\lambda_{\bm{x}}) = \frac{\sum\limits_{i=0}^m s_i(G^\lambda_{\bm{x}})}{m}
\end{equation}

We compute $s_{avg}$ for each image in $U_{val}$ and take the mean and standard deviation across all observations as

\begin{equation}
\mu^m_{val} = \frac{\sum\limits_{\bm{x} \in U_{val}} s_{avg}(G^\lambda_{\bm{x}})}{|U_{val}|}
\end{equation}

\begin{equation}
\sigma^m_{val} = \sqrt{\frac{\sum\limits_{\bm{x} \in U_{val}} (s_{avg}(G^\lambda_{\bm{x}}) - \mu_{val})^2}{| U_{val} | - 1}}
\end{equation}

We use $\mu^m_{val}$ and $\sigma^m_{val}$ to predict whether an input is adversarial or not. Specifically, if for a given input $\bm{x}$ we find that $s_{avg}(G_{\bm{x}}) > \mu^m_{val} + \sigma^m_{val}$, we flag it as an adversarial input. The results of this process are shown in Table \ref{table:average_match}. 

\subsubsection{Edge Counting and Average Edge Weight}

\begin{wrapfigure}{r}{0.35\linewidth}
\centering
  \includegraphics[width=\linewidth]{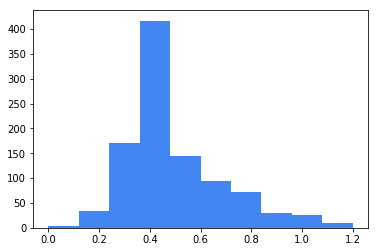}
  \includegraphics[width=\linewidth]{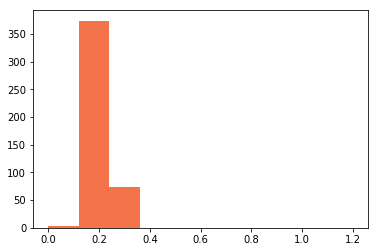}
\caption{\textit{Top}: Distribution of average edge weight for non-adversarial images. \textit{Bottom}: Distribution of average edge weight for adversarial images.}
\label{fig:edge_weight_distribution}
\end{wrapfigure}

\begin{figure}
\begin{subfigure}{.33\textwidth}
\centering
  \includegraphics[width=.75\linewidth]{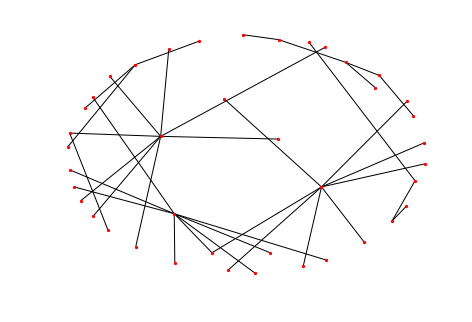} \vspace{5mm} \\
  \includegraphics[width=.75\linewidth]{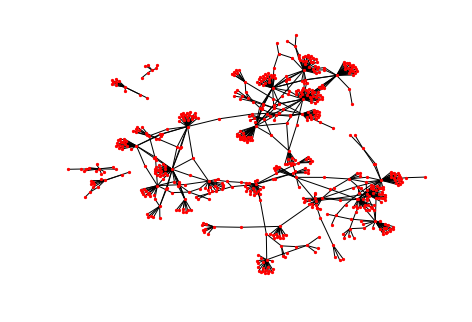}
\end{subfigure}
\begin{subfigure}{.33\textwidth}
\centering
  \includegraphics[width=.75\linewidth]{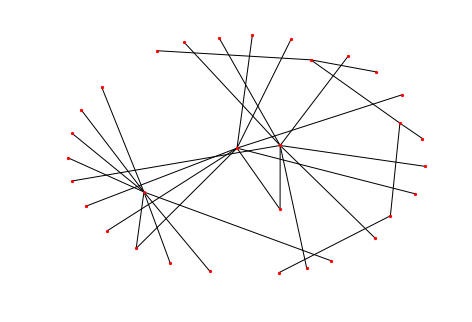} \vspace{5mm} \\
  \includegraphics[width=.75\linewidth]{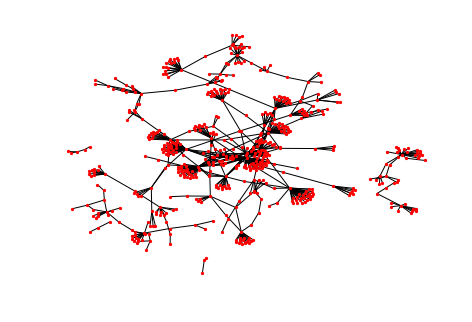}
 \end{subfigure}
\begin{subfigure}{.33\textwidth}
\centering
  \includegraphics[width=.75\linewidth]{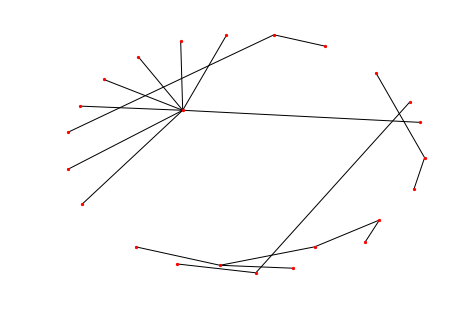} \vspace{5mm} \\
  \includegraphics[width=.75\linewidth]{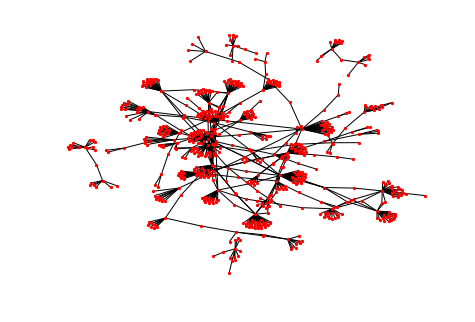}
 \end{subfigure}
\caption{Force-directed subgraphs from all components with lifetime > 0.1. \textit{Top}: Unperturbed MNIST images. \textit{Bottom}: Adversarial images with target class same as the graph above.}
\label{fig:subgraphs}
\end{figure}

Figure \ref{fig:subgraphs} shows an interesting property difference between adversarial and non-adversarial examples with respect to the induced persistent subgraph. It is clear that the persistent subgraphs induced by adversarial inputs contain many more edges compared to non-adversarial subgraphs for $\lambda = 0.1$. This property only appears for values near $\lambda = 0.1$ for our model trained on MNIST. For $\lambda < 0.1$, the edge distributions between adversarial and non-adversarial inputs begin to converge. For $\lambda > 0.1$, the edge count for non-adversarial inputs drops to zero. This fact implies there is an underlying difference in distribution of subgraph component lifetimes between adversarial and non-adversarial images. In particular, adversarial images induces more, longer-lived subgraphs than their non-adversarial counterparts. Perhaps this is an artifact of the optimization process used to generate the adversarial images wherein the optimization targets more robust (longer-lived) semantic subgraphs across classes. This is an interesting observation that warrants further investigation.

In conjunction with the difference in the induced number of edges between adversarial and non-adversarial images, we also observe that the average edge weights of the induced persistent subgraphs differ between the two types of inputs. Figure \ref{fig:edge_weight_distribution} shows how the average edge weight amongst persistent subgraphs induced by adversarial images is both lower and less variable than those induced by non-adversarial images. Again, this is an interesting observation that warrants further investigation. 

Regardless of the origin of these differences between adversarial and non-adversarial induced subgraphs, we can leverage these differences in order to detect adversarial inputs in our network. We follow a similar construction as in Section \ref{sec:max_node_matching} wherein we compute a statistic from $U_{train}$ and use this statistic to detect adversaries. 

\begin{wrapfigure}{l}{0.35\linewidth}
\centering
  \includegraphics[width=\linewidth]{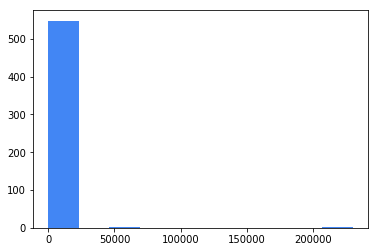}
  \includegraphics[width=\linewidth]{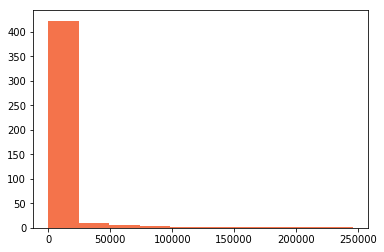}
\caption{\textit{Top}: Distribution of edge counts for non-adversarial images. \textit{Bottom}: Distribution of edge counts for adversarial images.}
\label{fig:edge_distributions}
\end{wrapfigure}

For differences in edge count, we first note that the distribution on the number of edges of the input-induced persistent subgraphs for both adversarial and non-adversarial inputs is highly skewed (Figure \ref{fig:edge_distributions}). We define $\mathscr{E}_{train} = \{ | E_{\bm{x}} | \ | \ E_{\bm{x}} \subset G^\lambda_{\bm{x}}, \bm{x} \in U_{train} \}$ as the edge counts of all persistent subgraphs induced by the training images. We then compute the median of this set of edge counts $m_{train} = \text{median}(\mathscr{E}_{train})$. We also compute the $\pi^{\text{th}}$ percentile of edge counts in $\mathscr{E}_{train}$. We denote the value of this percentile $p_{train}$. 

We use $m_{train}$ and $p_{train}$ to detect adversarial inputs in the neural network. For a given input $\bm{x}$, if $| E_{\bm{x}} | > m_{train} + p_{train}$ where $E_{\bm{x}} \subset G^\lambda_{\bm{x}}$, we flag $\bm{x}$ as an adversarial input. The results of this method are shown in Table \ref{table:edge_counts}. 

For differences in average edge weight, recall the definition of the persistent subgraph induced by an input $\bm{x}$ is $G^\lambda_{\bm{x}} = (V_{\bm{x}}, E_{\bm{x}}, \phi')$. We define the average edge weight of a persistent subgraph induced by $\bm{x}$ in an expected manner as

\begin{equation*}
	r_{avg}(G^\lambda_{\bm{x}}) = \frac{\sum\limits_{e \in E_{\bm{x}} \subset G^\lambda_{\bm{x}}} \phi'(e)}{| E_{\bm{x}} |}
\end{equation*}

We then calculate the average edge weight and standard deviation over the training examples as:

\begin{equation*}
	\mu^\phi_{train} = \frac{\sum\limits_{\bm{x} \in U_{train}} r_{avg}(G^\lambda_{\bm{x}})}{| U_{train} |}
\end{equation*}

\begin{equation*}
\sigma^\phi_{train} = \sqrt{\frac{\sum\limits_{\bm{x} \in U_{train}} (r_{avg}(G^\lambda_{\bm{x}}) - \mu^\phi_{train})^2}{| U_{train} | - 1}}
\end{equation*}

We use $\mu^\phi_{train}$ and $\sigma^\phi_{train}$ to determine whether a given input $\bm{x}$ is adversarial or not. Specifically, we flag an input $\bm{x}$ as adversarial if $r_{avg}(G^\lambda_{\bm{x}}) > \mu^\phi_{train} + \sigma^\phi_{train}$. The results of this approach are presented in Table \ref{table:average_weight}.

\section{Results}

We test each of the adversary detection algorithms described in the previous section. For each algorithm, we construct an image set of size 900 with 450 adversarial images (constructed as in Section \ref{sec:adversaries}) and 450 unaltered images. We hold $\rho = 0.99$ and test the detection capabilities across varying levels of hyperparameter $\lambda$ with $\kappa = 0$. We also test the detection capabilities for the highest and lowest values of $\lambda$ with the adversarial set created by setting $\kappa = 20$. We compute the accuracy and $\text{F}_1$-score for each test. We also report the false-positive and false-negative rate. Of special interest for our usage is the false-negative rate as this corresponds to the number of adversarial images that are not detected by the algorithm. 

\begin{table}[]
\centering
\caption{Adversary detection results for maximum node matching. The top results for both $\kappa$ values are in bold.}
\label{table:max_match}
\begin{tabular}{llllll}
$\kappa$ & $\lambda$ & Accuracy & False Positives & False Negatives & $\text{F}_1$-score \\
\cline{1-6}
0      & 0.1     & 0.683    & 235            & 50              & 0.737     \\
0      & 0.05    & 0.786    & 169            & 23              & 0.816     \\
0      & 0.01    & 0.858    & 109            & \textbf{19}              & 0.871     \\
0      & 0.001   & \textbf{0.864}    & \textbf{102}            & 20              & \textbf{0.876}     \\
20     & 0.1     & \textbf{0.925}    & \textbf{61}             & \textbf{6}               & \textbf{0.93}      \\
20     & 0.001   & 0.854    & 102            & 29              & 0.865    
\end{tabular}
\end{table}

Overall, comparing the average edge count of adversarially-induced persistent subgraphs versus the expected number of edges in the persistent subgraphs of non-adversarial inputs is the most effective detection algorithm for low-confidence adversaries ($\kappa = 0$). This detection mechanism is also the most effective for high-confidence adversaries ($\kappa = 20$). However, the maximum node matching approach only misses 6 adversaries which is the next best performing algorithm for $\kappa = 20$. 

\begin{table}[]
\centering
\caption{Adversary detection results for average node matching. The top results for both $\kappa$ values are in bold.}
\label{table:average_match}
\begin{tabular}{llllll}
$\kappa$ & $\lambda$ & Accuracy & False Positives & False Negatives & $\text{F}_1$-score \\
\cline{1-6}
0      & 0.1     & \textbf{0.885}    & \textbf{45}             & \textbf{58}              & \textbf{0.884}     \\
0      & 0.05    & 0.478    & 52             & 418             & 0.12      \\
0      & 0.01    & 0.476    & 57             & 414             & 0.133     \\
0      & 0.001   & 0.481    & 54             & 413             & 0.137     \\
20     & 0.1     & \textbf{0.938}    & \textbf{45}             & \textbf{11}              & \textbf{0.94}      \\
20     & 0.001   & 0.464    & 54             & 428             & 0.084    
\end{tabular}
\end{table}

\begin{table}[]
\centering
\caption{Adversary detection results for average edge weighting. The top results for both $\kappa$ values are in bold.}
\label{table:average_weight}
\begin{tabular}{llllll}
$\kappa$ & $\lambda$ & Accuracy & False Positives & False Negatives & $\text{F}_1$-score \\
\cline{1-6}
0      & 0.1     & 0.914    & \textbf{30}             & 47              & 0.913     \\
0      & 0.05    & \textbf{0.931}    & 33             & \textbf{29}              & \textbf{0.931}     \\
0      & 0.01    & 0.857    & 55             & 74              & 0.854     \\
0      & 0.001   & 0.788    & 122            & 69              & 0.774     \\
20     & 0.1     & \textbf{0.95}     & \textbf{30}             & \textbf{15}              & \textbf{0.951}     \\
20     & 0.001   & 0.865    & 69             & 52              & 0.868    
\end{tabular}
\end{table}

\begin{table}[]
\centering
\caption{Adversary detection results for average edge counting. The top results for both $\kappa$ values are in bold.}
\label{table:edge_counts}
\begin{tabular}{lllllll}
$\kappa$ & $\lambda$ & $\pi$ & Accuracy & False Positives & False Negatives & $\text{F}_1$-score  \\
\cline{1-7}
0      & 0.1   & 0.9  & \textbf{0.969}    & \textbf{18}             & \textbf{10}              & \textbf{0.969}   \\
0      & 0.05  & 0.9  & 0.816    & 7              & 158             & 0.78        \\
0      & 0.01  & 0.9  & 0.786    & 6              & 168             & 0.733      \\
20     & 0.1  & 0.95   & \textbf{0.984}    & \textbf{14}             & \textbf{0}               & \textbf{0.985}  
\end{tabular}
\end{table}

\section{Discussion and Future Work}

The results presented in this paper have shown numerous ways in which adversarial inputs may be detected based on the topological signature induced in the neural network. However, there are undoubtedly more and likely better ways to detect adversarial examples from their topological signatures. For example, the authors have experimented with edge matching with construction similar to the ranked node matching described in Section \ref{sec:max_node_matching}. However, the accuracy of this approach was consistently lower than that of maximum and average node matching. Our similarity measures are relatively simple, so future work may be in constructing more sophisticated similarity measures that are more sensitive to adversarial inputs. 

The primary downside of our topological adversary detection framework is in the computational cost. For each input, we compute the persistent homology on graphs with tens to hundreds of thousands, even with $\rho = 0.99$. As well, node matching is at least linear in the number of nodes of the computed persistent subgraphs. If the number of classes increases, the computational complexity of the maximum and average node matching algorithms increases as well. For a dataset with thousands of classes, these methods may prove computationally intractable.

Despite the computational complexity, our topological approach is robust. As we have seen, it is generally the case that these algorithms perform better with higher adversary distortion. As well, the authors cannot envision a white-box attack that would be able to significantly alter the performance of these algorithms given their deep integration with the features present in the input space.

Extending these ideas to other datasets and other network architectures is a natural source of future work. Due to the complexity constraints, the specific algorithms given above may not generalize well to larger datasets like CIFAR. However, a topological approach to detecting adversarial examples should still be fruitful. In fact, recent work \cite{su2017one} has shown that creating adversaries in more complex datasets like CIFAR is significantly easier than creating MNIST adversaries. It is reasonable to expect that, since the input CIFAR images are almost identical to their non-adversarial counterparts, we may be able to view this lack of distortion topologically. As well, many of the impediments to easily detecting adversarial MNIST images was in their homogeneity across classes. For example, eights and zeros contain highly similar pixel distributions. In a dataset with more heterogeneity across classes, the topological differences between classes may be more apparent, making class changes easier to detect. 

\section{Conclusion}

We have presented a method for topological analysis of neural network computations. By computing the persistent homology of the graph induced by input images, we constructed three robust algorithms for detection of adversarial inputs based on the topological information contained in the input-induced graph. Using images from MNIST, we show that our topological adversary detection method is able to detect adversaries with upwards of 98\% accuracy and $\text{F}_1$-score 0.98.

\bibliographystyle{plain}
\bibliography{adversary_detection_via_persistent_homology}

\end{document}